\documentclass{bmvc2k}

\title{MoG-VLM: Efficient Video Anomaly Detection through Motion-Guided Vision-Language Models}

\addauthor{Khaled Waleed Dawoud
}{khaled.dawoud@mbzuai.ac.ae}{1}
\addauthor{Muhammad Zaigham Zaheer}{Zaigham.Zaheer@mbzuai.ac.ae}{1}
\addauthor{Khan Muhammad}{khanmuhammad@skku.edu}{2}
\addauthor{Muhammad Haris Khan}{muhammad.haris@mbzuai.ac.ae}{1}

\addinstitution{
 Mohamed bin Zayed University of Artificial Intelligence,\\
 Abu Dhabi,\\
 United Arab Emirates
}
\addinstitution{
Sungkyunkwan University,\\
Seoul 03063,\\
South Korea
}

\runninghead{Dawoud, ET AL}{MoG-VLM}

\def\etal{\emph{et al}\bmvaOneDot}

\usepackage[utf8]{inputenc}
\usepackage{amsmath,amssymb}
\usepackage{graphicx}
\usepackage{booktabs}
\usepackage{tabularx}
\usepackage{multirow}
\usepackage{textgreek}
\usepackage{pifont}      
\usepackage{float}
\usepackage{enumitem}
\usepackage{caption}
\usepackage{url}
\usepackage[table]{xcolor}
\usepackage[most]{tcolorbox}   

\definecolor{framecol}{RGB}{255,242,232}
\definecolor{temporalcol}{RGB}{235,245,255}
\definecolor{dualcol}{RGB}{232,246,239}
\definecolor{refinecol}{RGB}{245,245,245}
\definecolor{eqten}{RGB}{0,102,153}
\definecolor{captionblue}{RGB}{0,51,102}
\definecolor{eqeleven}{RGB}{102,51,153}

\begin{document}

\maketitle

\begin{abstract}
Vision-Language Models (VLMs) have shown strong potential for video anomaly detection, but dense frame-level inference makes them impractical for real-time deployment. Existing acceleration commonly relies on uniform sampling, which ignores content dynamics and can either waste computation on static segments or miss localized anomalies.
We propose MoG-VLM a motion-adaptive anomaly detection framework that leverages motion purely as a lightweight control signal to decide when VLM inference is necessary. To this end, we introduce Temporal Gradient Magnitude (TGM) which performs content-aware sampling to concentrate computation on informative regions while avoiding redundant processing.
We further introduce a hierarchical dual-stream design in which dense multi-frame VLM reasoning is activated only around frames with elevated anomaly likelihood, via anchor-triggered context-window expansion with early termination. A fusion module then reconciles these instantaneous semantic cues with temporal action dynamics.
Across four benchmarks, our method achieves 91.91\%, 84.30\%, 87.88\%, and 78.96\% AUC-ROC on XD-Violence, UCF-Crime, MSAD, and ShanghaiTech, respectively, establishing new state of the art on three benchmarks under the zero-shot, training-free setting. On XD-Violence, we improve over the previous best result by $+$0.57\% while delivering a 1.31$\times$ inference speedup. These results demonstrate that content-aware allocation of VLM capacity, processing fewer frames more intelligently rather than all frames uniformly, is a more effective axis of progress than uniformly densifying VLM inference across frames. Code is available on GitHub:  
\url{https://github.com/kdawoud91/MoG-VLM}.

\end{abstract}

\section{INTRODUCTION}
The proliferation of camera systems has generated unprecedented volumes of video data, creating an urgent need for automated anomaly detection systems capable of identifying rare and unpredictable events in real-time \cite{sultani2018real,dawoud2025fusedvision}. 
Traditional approaches~\cite{dawoud2025fusedvision,zanella2024harnessing,lin2025unified} that process every frame with equal computational intensity or uniform stride are fundamentally misaligned with the nature of video footage, where anomalous events are sparse and localized. This inefficiency becomes particularly pronounced when deploying computationally expensive Vision-Language Models (VLMs), which offer unprecedented semantic understanding but impose prohibitive inference costs for continuous real-time processing \cite{lin2024video,Qwen-VL}.
Modern video-capable architectures such as Video-LLaMA \cite{zhang2023video} and the Qwen-VL family \cite{Qwen-VL} demonstrate remarkable improvements in temporal coherence and semantic depth compared to earlier contrastive models \cite{radford2021learning,alayrac2022flamingo}, making them particularly well-suited for 
applications where abnormality is often defined by context. However, this representational power comes at a substantial computational cost. Without mechanisms to adapt inference effort to content dynamics in video anomaly detection (VAD), this representational power remains largely inaccessible under the computational and memory constraints of real-time deployment~\cite{zanella2024harnessing}.

\begin{figure}[t]
    \centering
    \captionsetup{skip=4pt,labelfont={color=captionblue},textfont={color=captionblue}}
    \includegraphics[width=\textwidth]{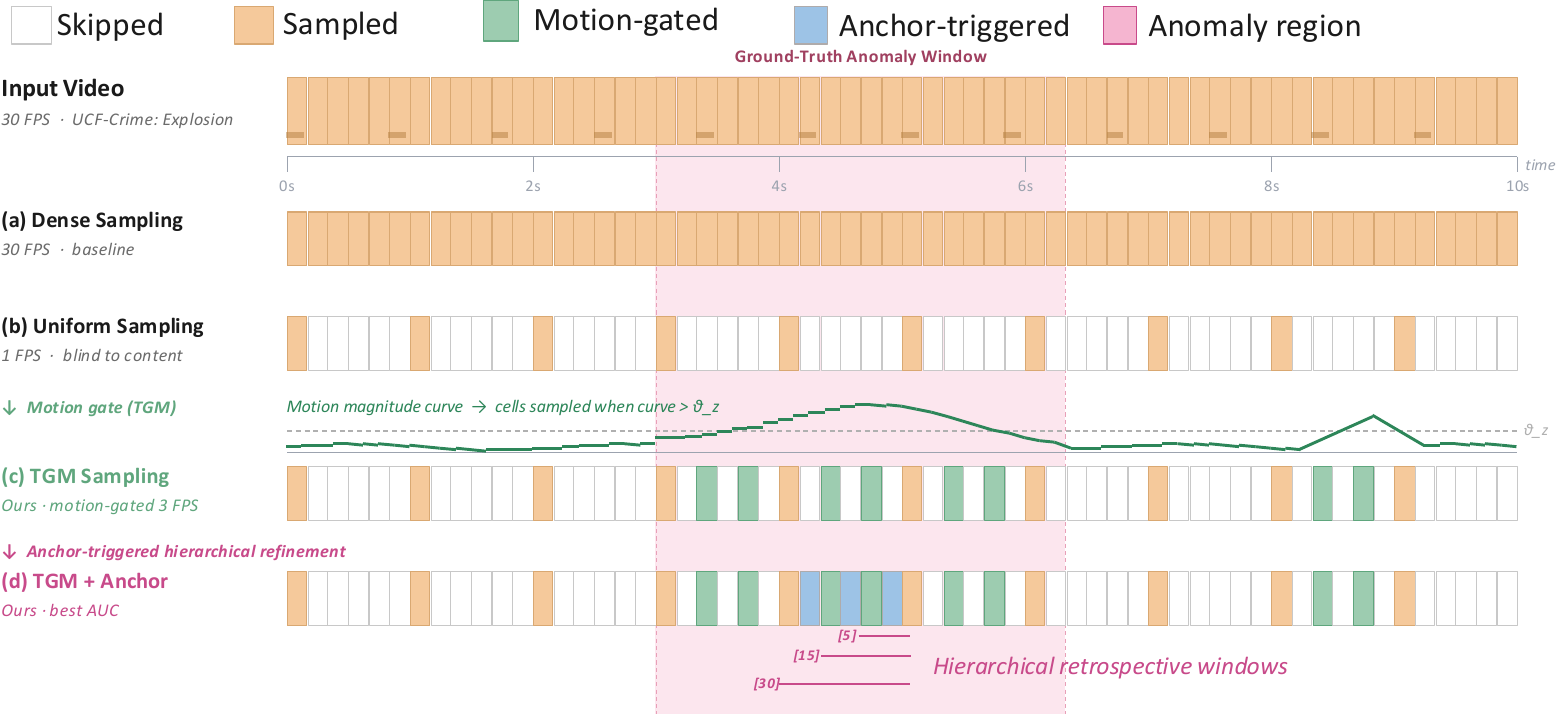}
    \caption{
\textbf{Comparison of frame sampling strategies for video anomaly detection.}
(a) Dense sampling, incurring prohibitively high computational cost~\cite{shao2025eventvad}.
(b) Uniform sampling, which can waste computation on static segments~\cite{zanella2024harnessing, lin2025unified}.
(c) Our motion-adaptive sampling method based on Temporal Gradient Magnitude (TGM) dynamically selects frames according to motion patterns, reducing processed frames.
(d) Our introduced anchor-triggered processing selectively expands time windows around high-confidence detections.
}
    \label{fig:Overview}
    \vspace{-0.5em}
\end{figure}

Current VLM-based anomaly detection methods overwhelmingly adopt uniform processing paradigms, sampling frames at fixed strides ~\cite{zanella2024harnessing, lin2025unified, ye2025vera} or analyzing entire videos exhaustively~\cite{shao2025eventvad} regardless of content dynamics. This is wasteful: many video intervals are static or low-activity yet incur the same per-frame VLM cost as dynamic, anomaly-rich segments. Recent zero-shot extensions~\cite{zanella2024harnessing, lin2025unified} and parameter-efficient fine-tuning approaches~\cite{ye2025vera, yang2024text} have advanced reasoning quality, but inherit this uniform-sampling assumption and overlook the spatial and temporal sparsity inherent to surveillance footage.

In contrast, our framework operates from a deliberately coarse baseline sampling rate and selectively densifies only where motion dynamics indicate informative content. As illustrated in Fig.~\ref{fig:Overview}, dense (Fig.~\ref{fig:Overview}a) and uniform (Fig.~\ref{fig:Overview}b) sampling process frames indiscriminately, while our motion-adaptive approach (Fig.~\ref{fig:Overview}c-d) concentrates computation on temporally dynamic segments intervals exhibiting high or sudden motion changes characteristic of fights, explosions, or vehicle collisions and suppresses redundant inference on static footage.

This gap between semantic capability and computational tractability motivates a fundamental question: \textit{Starting from a coarse uniform sampling baseline, can we adaptively allocate additional computation only when video dynamics demand it to achieve real-time VLM-based anomaly detection?} Addressing this question requires moving beyond accuracy-centric evaluation toward a holistic framework that explicitly balances detection performance with computational efficiency. Specifically, there exists no prior work that systematically investigates adaptive mechanisms for content-aware resource allocation in vision-language-based anomaly detection, nor sufficient analysis of the trade-offs between computational cost and video anomaly detection performance.

In this work, we introduce a motion-adaptive anomaly detection framework that fundamentally reimagines how computational resources are allocated across video content. Unlike classical approaches that employ motion as a direct classification feature ~\cite{feichtenhofer2016convolutional, scovanner20073, park2020learning, georgescu2021anomaly}, we take advantage of motion purely as a lightweight control mechanism, a semantic activation signal that determines when  VLM inference is necessary. We introduce TGM analysis to establish a motion quantification mechanism, thereby dynamically adjusting the sampling density based on detected temporal dynamics.
We further introduce an anchor-triggered hierarchical refinement mechanism that preserves computational efficiency while recovering fine-grained temporal context. Adaptively sampled frames that exhibit high anomaly confidence serve as anchor points, triggering progressive multi-window analysis, ensuring that dense coverage is applied exclusively where anomalous patterns emerge.
In parallel, we introduce a complementary temporal stream that captures action dynamics through video-language modeling. This branch processes non-overlapping temporal windows, generating video-level semantic descriptions that encode motion trajectories and behavioral sequences. 
The two pathways converge through adaptive scoring, combining frame-level instantaneous detections with temporal-level behavioral context into unified predictions. 

\noindent\textbf{Contributions.} We summarize our key contributions as follows: \textbf{Motion-Adaptive Sampling Framework:} A novel motion quantification mechanism using multiscale TGM analysis and adaptive z-score that dynamically adjusts sampling density based on temporal dynamics, achieving substantial computational savings. \textbf{Anchor-Triggered Hierarchical Refinement:} A progressive multi-window expansion mechanism where adaptively sampled frames exhibiting high anomaly trigger conditional retrospective analysis, examining preceding context with early termination to prevent unnecessary computation on low-confidence detections. \textbf{Tier-Aware Adaptive Fusion:} A fusion mechanism in which the mixing weight λ between frame-level and temporal-level streams is not learned but inherited from the TGM motion tier of each frame, with λ taking a piecewise-constant value determined by whether the frame is unflagged, motion-flagged, or anchor-triggered. This tier-driven weighting is agnostic to dataset specifics and scene dynamics. \textbf{Comprehensive Efficiency-Accuracy Analysis:} A systematic study of the compute-accuracy trade-off in zero-shot VLM-based anomaly detection, a dimension largely absent from prior work. MoG-VLM attains 84.30\%, 91.91\%, 87.88\% and 78.96\% AUC-ROC on UCF-Crime, XD-Violence, MSAD and Shang setting a new zero-shot training-free state of the art on three benchmarks.

\section{Related Work}
Video anomaly detection (VAD) aims to temporally localise frames containing events that deviate from expected behaviour. Early supervised methods required frame-level annotations~\cite{sultani2018real}, while weakly-supervised formulations reduced the annotation burden to video-level labels through multiple-instance learning (MIL)~\cite{sultani2018real,Chen,Learning_Local_Semantic,Wang_2026_CVPR,wang2025learning}. The emergence of large vision–language models (VLMs) and large language models (LLMs) has opened a new paradigm for VAD by inferring anomaly scores from natural-language scene descriptions at test time, without gradient-based training. LAVAD~\cite{zanella2024harnessing} pioneered this direction by captioning each frame with a frozen VLM, summarising temporal windows with an LLM, and refining anomaly scores through cross-modal similarity. Despite strong results, its uniform processing of every sampled frame regardless of scene content leads to 71.4 GPU-hours on the XD-Violence test set alone, allocating identical computation to static backgrounds and fast-evolving anomalous events alike. Subsequent work has focused on reasoning quality rather than efficiency. \textit{Unified-reasoning}~\cite{lin2025unified} introduces a score-based gate that triggers a second LLM pass with automatically extracted anomaly-tag priors for ambiguous videos, raising AUC-ROC to 84.28\% on UCF-Crime and 91.34\% on XD. Yet the underlying sampling stays content-blind, a fixed stride of 16 frames is applied uniformly, expending VLM inference on static footage
regardless of scene dynamics.
Other VLM-based methods relax the strictly zero-shot constraint. VERA~\cite{ye2025vera} steers a frozen VLM with verbalized guiding questions optimised from labelled data, without modifying model weights, though the prompt optimisation itself requires an annotated training set. Holmes-VAU~\cite{zhang2025holmes} and Hawk~\cite{tang2024hawk} instruction-tune multimodal LLMs on anomaly clips to produce rich explanations, but they remain tied to seen anomaly types and incur high annotation costs. STPrompt~\cite{wu2024weakly} learns spatio-temporal prompts under weak class-level supervision. AnyAnomaly~\cite{ahn2026anyanomaly} proposes customisable VAD by casting detection as context-aware visual question answering with a frozen LVLM, and EventVAD~\cite{shao2025eventvad} combines graph architectures with multimodal LLMs through temporal reasoning. 
The literature has overwhelmingly prioritised detection accuracy~\cite{ahn2026anyanomaly,shao2025eventvad,zhou2023dual} over computational efficiency, leaving the deployment feasibility of
VLM-based VAD unclear. Among the methods discussed above, only \textit{Unified}~\cite{lin2025unified} reports timing information. This gap is significant: real-world video systems must process hours of continuous footage, and the practical feasibility of VLM-based anomaly detection depends critically on throughput, not solely on AUC-ROC.
The principle of allocating computation only where content demands it has a long history in video understanding. In action recognition, AdaFrame~\cite{wu2019adaframe} trains a policy network to adaptively select informative frames, SCSampler~\cite{korbar2019scsampler} routes only salient clips to a heavy 3D backbone, AR-Net~\cite{meng2020ar} learns per-frame resolution selection, FrameExit~\cite{ghodrati2021frameexit} conditions early-exit decisions on intermediate features, and LiteEval~\cite{wu2019liteeval} alternates between a coarse and fine network via a gating module. The shared design pattern is a two-stage architecture: a cheap first pass produces a saliency signal, and a more expensive pass is activated selectively.
Our framework inherits this two-stage philosophy but differs in three respects. First, our gating mechanism is entirely training-free, we threshold a multiscale Temporal Gradient Magnitude (TGM) via adaptive z-scores, requiring no learned parameters, whereas the methods above learn sampling policies from labelled data, an impractical requirement in anomaly detection where positive examples are scarce. Second, we target untrimmed footage spanning minutes to hours rather than trimmed classification clips, where the informative-to-static ratio is far more extreme. Third, motion serves purely as a \emph{control signal} that decides \emph{when} VLM inference is invoked, it does not enter the anomaly score itself, avoiding the conflation of motion salience with anomaly likelihood.

\begin{figure*}[t]
\centering
\captionsetup{skip=4pt,labelfont={color=captionblue},textfont={color=captionblue}}
\includegraphics[width=\textwidth,height=0.4\textheight,keepaspectratio]{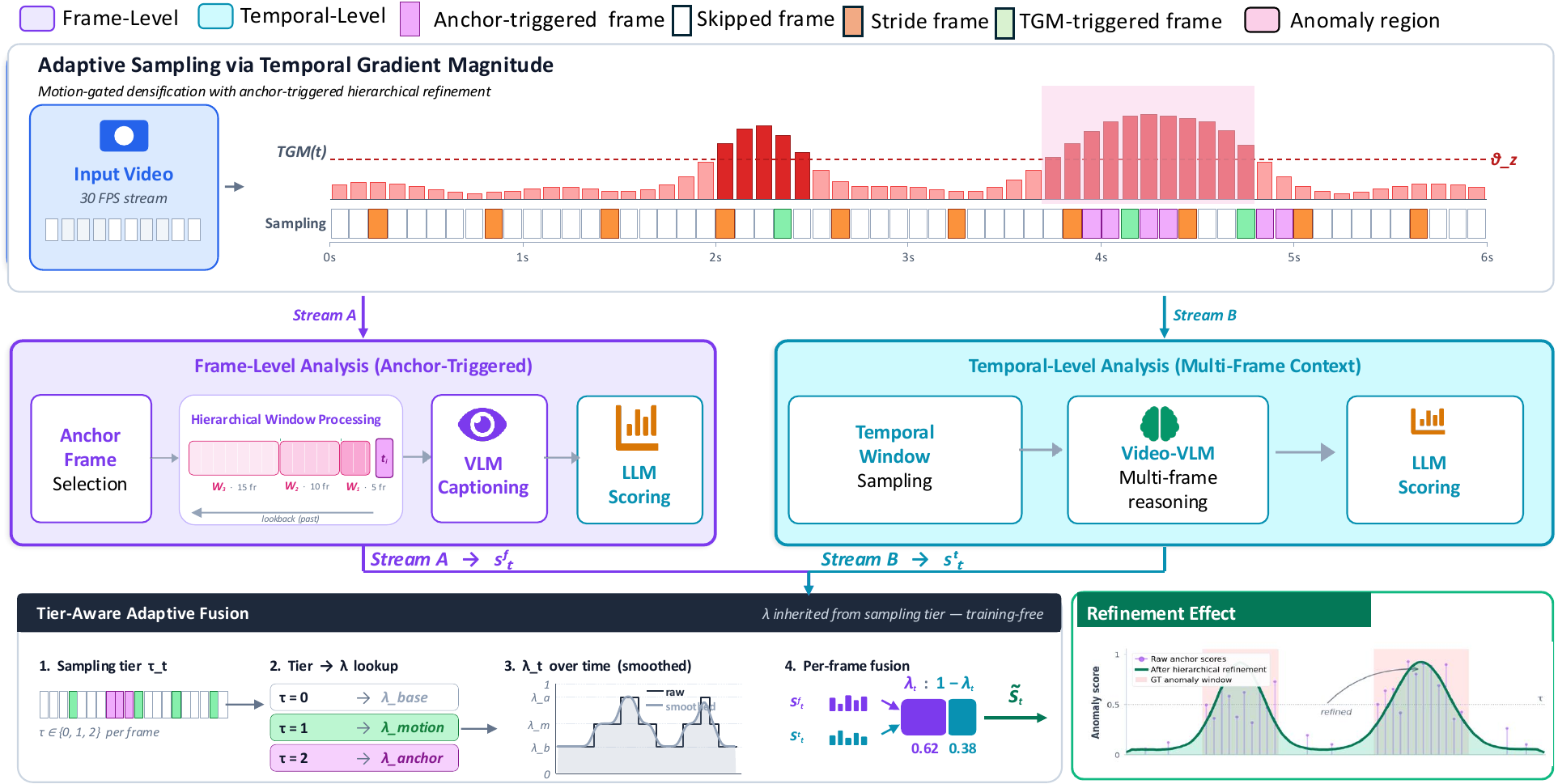}
\caption{Overview of the MoG-VLM framework. Motion-adaptive sampling via TGM categorizes frames into three tiers: baseline uniform (orange), TGM-triggered densification (green), and anchor-triggered (magenta). 
Baseline stride frames and TGM-triggered frames with high anomaly scores become anchors, initiating hierarchical lookback windows ($W_1$, $W_2$, $W_3$). Stream~B conducts temporal-level Video-VLM multi-frame reasoning and LLM scoring. Tier-aware adaptive fusion combines both streams using motion-tier-inherited weights ($\lambda_{\text{base}}$, $\lambda_{\text{motion}}$, $\lambda_{\text{anchor}}$) without training.}
\label{fig:architecture}
\end{figure*}

\section{Methodology}
\label{sec:methodology}
Our framework introduces motion-adaptive sampling with hierarchical VLM-based processing through a dual-stream architecture that balances computational efficiency with detection accuracy. Fig.~\ref{fig:architecture} illustrates the complete pipeline.

\subsection{\textbf{Frame-Level Anomaly Detection via Motion-Adaptive Sampling}}
\label{sec:frame_level}
We propose a motion-adaptive sampling strategy that dynamically adjusts processing density based on temporal dynamics, enabling efficient allocation of computational resources to informative regions while maintaining detection accuracy.

\subsubsection{Temporal Gradient Magnitude Analysis}
\label{sec:tgm_analysis}
Given a video sequence $V = \{I_t\}_{t=1}^{T}$, we extract frames at a baseline sampling rate to establish coarse temporal coverage. For each sampled frame $I_t$, we compute the Temporal Gradient Magnitude (TGM) at multiple temporal scales $\mathcal{W} = \{w_1, \ldots, w_K\}$ to capture motion patterns ranging from instantaneous changes to sustained activities. At each scale $w$, we form per-pixel velocity $v_w = (I_t - I_{t-w})/w$ and acceleration $a_w = (I_t - 2I_{t-w} + I_{t-2w})/w^2$, and reduce the resulting TGM map to a scalar via its $90^{\text{th}}$ percentile:
\begin{equation}
m_w(t) = \mathrm{P}_{90}\!\left[\sqrt{v_w^2 + \beta\, a_w^2}\,\right],
\end{equation}
where $\beta = 0.35$ controls the relative contribution of acceleration to the combined motion signal. The per-scale summaries are aggregated by magnitude-weighted averaging,
\begin{equation}
\text{TGM}(t) = \sum_{w \in \mathcal{W}} \tilde{\alpha}_w(t)\, m_w(t), 
\quad 
\tilde{\alpha}_w(t) = \frac{m_w(t)}{\sum_{w' \in \mathcal{W}} m_{w'}(t)}.
\end{equation}
Scales exhibiting stronger response contribute proportionally more to the aggregated signal.

\subsubsection{Adaptive Z-Score Anomaly Detection}
\label{sec:adaptive_zscore}
To identify temporal regions exhibiting anomalous motion patterns, we employ adaptive z-score normalization over a sliding exponential moving average (EMA) of TGM statistics. This approach enables context-aware anomaly detection that adapts to scene-specific activity levels while remaining sensitive to deviations from normal behavior.

For each frame $t$, we compute the adaptive z-score:
\begin{equation}
z(t) = \frac{\text{TGM}(t) - \mu_t}{\sigma_t + \epsilon}
\end{equation}
where $\mu_t$ and $\sigma_t$ are the running mean and standard deviation maintained via EMA updates with decay factor $\alpha$. Critically, we implement a freezing mechanism during anomaly detection windows: when $z(t) > \theta_{\text{anom}}$ (where $\theta_{\text{anom}}$ is the anomaly threshold), we freeze updates to $\mu_t$ and $\sigma_t$ to prevent contamination of the reference distribution by anomalous behavior. This ensures that detected anomalies do not erroneously recalibrate, which would reduce sensitivity to sustained anomalous activities.

\subsubsection{Motion-Guided Adaptive Sampling}
\label{sec:adaptive_sampling}
Upon detecting motion anomalies via z-score thresholding, we trigger adaptive refinement of the temporal sampling rate in the identified region. Specifically, when $z(t) > \theta_{\text{anom}}$, we extend the detection window to $[t-\Delta t, t+\Delta t]$ and increase the sampling rate to $r_{\text{high}}$ within this temporal neighborhood, while maintaining the baseline rate $r_{\text{base}}$ elsewhere. This produces a non-uniform sampling schedule:
\begin{equation}
\label{eq:TQM}
r(t) = \begin{cases} 
r_{\text{high}} & \text{if } \exists \tau: z(\tau) > \theta_{\text{anom}} \text{ and } |t - \tau| \leq \Delta t \\
r_{\text{base}} & \text{otherwise}.
\end{cases}
\end{equation}
This adaptive sampling strategy concentrates computational resources on temporally dynamic regions while avoiding redundant processing of static or slowly-changing segments.

\subsection{\textbf{Vision Language Frame-Level Captioning}}
\label{sec:vlm_captioning}
For each adaptively sampled frame, we employ a vision--language model to generate descriptions of the visual content.
Given a sampled frame $I_t$, frame-level captioning is formulated as:
\begin{equation}
\label{eq:vlm_caption}
c_t = \mathcal{F}_{\text{VLM}}(I_t; \mathcal{P}_{\text{frame}}).
\end{equation}
$\mathcal{F}_{\text{VLM}}$ denotes the vision--language model and $\mathcal{P}_{\text{frame}}$ is a task-specific prompt designed to emphasize anomaly-relevant attributes such as objects, persons, and spatial relationships. For exact prompts please see Supplementary Section C.3.1

\subsection{\textbf{Language-Based Anomaly Scoring}}
\label{sec:anomaly_scoring}
Given a frame caption $c_t$, we prompt the language model to assess the anomaly likelihood through structured reasoning. The prompt template $\mathcal{P}_{\text{score}}^{\text{frame}}$ is designed to emphasize instantaneous anomaly indicators, instructing the model to identify unusual actions, dangerous situations, suspicious behaviors, violence, accidents, or any visual patterns that deviate from typical video footage. (see Supplementary~C.3.2 for the full prompts)
The frame-level anomaly score at time $t$ is computed as:
\begin{equation}
\label{eq:frame_scoring}
s_t^{\text{frame}} = \mathcal{F}_{\text{LM}}(c_t;\, \mathcal{P}_{\text{score}}^{\text{frame}}).
\end{equation}
$\mathcal{F}_{\text{LM}}$ denotes the language model and $s_t^{\text{frame}} \in [0, 1]$ represents the frame-level anomaly score.

\subsection{\textbf{Anchor-Triggered Multi-Frame Analysis}}
\label{sec:motion_triggered_temporal}
While the frame-level analysis provides efficient anomaly detection through motion-adaptive sampling at an elevated rate $r_{\text{high}}$ in dynamic regions, certain anomalous activities require even denser temporal coverage to fully capture their fine-grained dynamics. Rather than indiscriminately escalating the sampling rate across all dynamic segments, we introduce an anchor-triggered multi-window mechanism that selectively activates dense multi-frame processing only around adaptively sampled frames that exhibit high anomaly likelihood.

\subsubsection{Anchor-Based Processing}
\label{sec:anchor_temporal}
From the motion-adaptive sampling pipeline (Section~\ref{sec:frame_level}), we obtain frame-level anomaly scores $s_t^{\text{frame}}$ at non-uniform positions determined by the adaptive sampling rate $r(t)$ (Eq.~\ref{eq:TQM}). The full set of adaptively sampled frames is $\mathcal{T}_{\text{sample}} = \mathcal{T}_{\text{base}} \cup \mathcal{T}_{\text{motion}}$, where $\mathcal{T}_{\text{base}}$ contains frames sampled at the baseline stride rate $r_{\text{base}}$ and $\mathcal{T}_{\text{motion}}$ contains additional frames densified at $r_{\text{high}}$ within motion-gated regions. Each frame $t_i \in \mathcal{T}_{\text{sample}}$ serves as a potential anchor for triggering fine-grained window analysis.

A frame $t_i \in \mathcal{T}_{\text{sample}}$ qualifies as a trigger anchor when its frame-level anomaly score exceeds an activation threshold $\theta_{\text{trigger}}$. We define the set of trigger anchors as:
\begin{equation}
\label{eq:anchor_set}
\mathcal{T}_{\text{anchor}} = \left\{ t_i \in \mathcal{T}_{\text{sample}} \;\middle|\; s_{t_i}^{\text{frame}} > \theta_{\text{trigger}} \right\}.
\end{equation}
Since $\mathcal{T}_{\text{sample}}$ inherits the non-uniform structure of $r(t)$, anchor candidates emerge naturally from motion dynamics rather than fixed intervals. Dynamic regions (where $r(t) = r_{\text{high}}$) yield denser anchor candidates, while static regions (where $r(t) = r_{\text{base}}$) yield anchor candidates at the baseline uniform sampling density. For each $t_i \in \mathcal{T}_{\text{anchor}}$, we initiate a hierarchical temporal analysis that progressively examines preceding frames at the native video frame rate to capture fine-grained dynamics.

\subsubsection{Hierarchical Window Processing}
\label{sec:hierarchical_windows}
For each anchor frame $t_i \in \mathcal{T}_{\text{anchor}}$, we apply a three-stage hierarchical processing strategy that progressively expands the amount of pre-anchor context examined at full resolution.
We define three non-overlapping lookback windows of increasing duration, parameterized by $0 < \Delta_1 < \Delta_2 < \Delta_3$:
$W_1(t_i)=[t_i-\Delta_1,t_i)$, $W_2(t_i)=[t_i-\Delta_2,t_i-\Delta_1)$, and $W_3(t_i)=[t_i-\Delta_3,t_i-\Delta_2)$.
At each stage $k\in\{1,2,3\}$, we run the same frame-level pipeline on frames in $W_k(t_i)$ to obtain scores $s_{\tau}^{\text{frame}}$, and compute the mean score over that window:

\begin{equation}
\label{eq:hierarchical_scoring}
\bar{s}_k(t_i) = \frac{1}{|W_k(t_i)|} \sum_{\tau \in W_k(t_i)} s_{\tau}^{\text{frame}}.
\end{equation}

Stage progression is conditioned on $\bar{s}_{k}(t_i) > \theta_{\text{trigger}}$ for $k \in \{1,2\}$, so additional high-resolution is only processed when earlier evidence suggests sustained anomalous activity.

\subsection{\textbf{Temporal-Level Anomaly Detection via Multi-Frame Context}}
\label{sec:temporal_level}
While frame-level analysis captures fine-grained semantic understanding of individual moments, many anomalous activities exhibit temporal dependencies spanning multiple frames. To capture these dynamics, we introduce a complementary temporal-level stream that processes multi-frame sequences through a video-language model.

\subsubsection{Temporal Window Construction}
\label{sec:temporal_windows}
We construct temporal windows at regular intervals along the video to capture local motion patterns and activity evolution. For a video sequence $V = \{I_t\}_{t=1}^{T}$, target frames are selected at a stride $s$, and around each target $t_{\text{target}}^{(i)}$ we form a window $\mathcal{W}_i$ of duration $2\omega$ seconds by sampling $n_{\text{frames}}$ frames:
\begin{equation}
\label{eq:temporal_window}
\mathcal{W}_i = \left\{ I_{t_j^{(i)}} \right\}_{j=1}^{n_{\text{frames}}}, \quad t_j^{(i)} \in \left[t_{\text{target}}^{(i)} - \omega,\; t_{\text{target}}^{(i)} + \omega\right],
\end{equation}
where the $n_{\text{frames}}$ sample positions are spaced across the interval. This ensures the video-language model receives a fixed-size input that spans the full window duration. 

\subsubsection{Video-Language Temporal Captioning and Scoring}
\label{sec:video_temporal}
For each window $\mathcal{W}_i$, we employ a video-language model that incorporates spatio-temporal attention to capture motion patterns, action progressions, and inter-frame dependencies across the sequence. The temporal caption is generated as:
\begin{equation}
\label{eq:temporal_caption}
c_i^{\text{temporal}} = \mathcal{F}_{\text{VideoLLM}}(\mathcal{W}_i;\, \mathcal{P}_{\text{temporal}}),
\end{equation}
where $\mathcal{F}_{\text{VideoLLM}}$ denotes the video-language model and $\mathcal{P}_{\text{temporal}}$ is a prompt template designed to elicit descriptions of activities, motion dynamics, and behavioral patterns. (see Supplementary~C.3.3 for the full prompts). Following the frame-level scoring procedure, the temporal anomaly score for window $i$ is then obtained via the same language model:
\begin{equation}
\label{eq:temporal_scoring}
s_i^{\text{temporal}} = \mathcal{F}_{\text{LM}}(c_i^{\text{temporal}};\, \mathcal{P}_{\text{score}}^{\text{temporal}}),
\end{equation}
with $s_i^{\text{temporal}} \in [0, 1]$. Window-level scores are propagated forward in time until the next window, so every frame in the evaluation timeline inherits the most recent window-level score, mirroring the forward-fill rule used for frame-level scores.

\subsection{Tier-Aware Adaptive Fusion}
\label{sec:adaptive-fusion}

The frame and temporal streams provide complementary evidence:
$s^{\mathrm{frame}}_t$ captures instantaneous semantics from adaptively sampled
frames, while $s^{\mathrm{temporal}}_t$ encodes action dynamics over multi-frame
windows. We fuse them per frame through a combination with a
time-varying weight $\lambda_t \in [0,1]$:
\begin{equation}
\label{eq:fusion}
\tilde{s}_t \;=\; \lambda_t\, s^{\mathrm{frame}}_t \;+\; (1-\lambda_t)\, s^{\mathrm{temporal}}_t .
\end{equation}

Rather than treating $\lambda_t$ as a global hyperparameter, we condition it
on the sampling tier $\tau_t \in \{0,1,2\}$ inherited from the
motion-adaptive pipeline: \emph{base} ($\tau_t{=}0$, baseline rate
$r_{\mathrm{base}}$), \emph{motion} ($\tau_t{=}1$, TGM-densified to
$r_{\mathrm{high}}$), and \emph{anchor} ($\tau_t{=}2$, hierarchical
refinement at full resolution). Tier membership maps to $\lambda_t$ through
three scalars:
\begin{equation}
\label{eq:lambda-tiers}
\lambda_t \;=\;
\begin{cases}
\lambda_{\mathrm{base}},   & \tau_t = 0,\\[2pt]
\lambda_{\mathrm{motion}}, & \tau_t = 1,\\[2pt]
\lambda_{\mathrm{anchor}}, & \tau_t = 2,
\end{cases}
\quad
\lambda_{\mathrm{base}} \,<\, \lambda_{\mathrm{motion}} \,<\, \lambda_{\mathrm{anchor}}.
\end{equation}

Each frame's fusion coefficient $\lambda_t$ corresponds to its tier, and trust in $s^{\mathrm{frame}}_t$ increases monotonically with tier level. Base-tier frames are uniformly sampled and receive the smallest coefficient, which shifts emphasis to the temporal stream. Motion-tier frames are densified by TGM and receive an intermediate coefficient. Anchor-tier frames are refined hierarchically, receive the largest coefficient, and dominate the fused score. A frame's tier reflects the highest level of processing it receives, baseline-stride and TGM-densified frames are upgraded to $\tau_t=2$ when they qualify as anchors (Eq.~\ref{eq:anchor_set}). We smooth $\lambda_t$ with a 1-D Gaussian and clip to $[0,1]$ before applying Eq.~\ref{eq:fusion} to avoid discontinuities at tier boundaries, then apply a second Gaussian to $\tilde{s}_t$ for evaluation stability. 

\section{Experiments}

\subsection{Datasets \& Evaluation Metrics}

\textbf{XD-Violence Dataset}~\cite{wu2020not}. XD-Violence (XD) is a large-scale anomaly detection benchmark that contains 800 videos for testing, covering six violence categories. \textbf{UCF-Crime Dataset}~\cite{sultani2018real}. UCF is a large-scale anomaly detection benchmark with 290 untrimmed videos for testing across 13 categories. \textbf{MSAD~\cite{zhu2024advancing}} is a recent multi-scenario benchmark offering 360 test videos across 14 scene types with varied camera perspectives.
\textbf{ShanghaiTech Dataset}~\cite{liu2018future}. ShanghaiTech contains 107 testing videos with abnormal events captured from fixed cameras. Given its scale and category diversity, we anchor our ablation and efficiency analysis on XD-Violence and report supporting numbers on the remaining benchmarks. 
\textbf{Evaluation Metrics.} We report frame-level AUC-ROC, the
standard threshold-independent metric for video anomaly
detection.

\subsection{Implementation Details}

\textbf{Motion-Adaptive Sampling.} We extract frames at a baseline rate of 1 FPS to establish coarse temporal coverage. For TGM computation, we employ multiscale temporal analysis with window sizes $\mathcal{W} = \{1, 2, 4\}$ frames to capture motion patterns at different temporal scales. Adaptive z-score anomaly detection uses an exponential moving average (EMA) decay factor of $\alpha = 0.95$ for baseline statistics tracking, with anomaly threshold $\theta_{\text{anom}} = 2.3$ standard deviations. 
When motion anomalies are detected, we increase sampling density to $r_{\text{high}} = 3$ FPS within a temporal window of $\Delta t = 1$ seconds around the detected frame. 
\textbf{Vision-Language Models.} We employ Qwen3-VL-8B-Instruct for frame-level captioning and VideoLLaMA3-7B for temporal-level processing, where each input consists of $n_{\text{frames}} = 10$ frames uniformly sampled from a non-overlapping 64-frame window at the video frame rate.
\textbf{Language-Based Anomaly Scoring.} For both frame-level and temporal-level, we employ Qwen3-4B-Instruct-2507. The scoring prompts employ chain-of-thought reasoning, instructing the model to first analyze the caption content, then assess anomaly likelihood on a scale from 0 (completely normal) to 1 (highly anomalous).
Run-to-run AUC variation across three reruns remains below 0.20\% on all benchmarks (variance attributable to LLM temperature sampling).
\textbf{Anchor-Triggered Multi-Frame Analysis.} When a frame's anomaly score exceeds $\theta_{\text{trigger}}{=}0.5$, we initiate hierarchical lookback windows of $[t_i{-}5, t_i)$, $[t_i{-}15, t_i{-}5)$, and $[t_i{-}30, t_i{-}15)$ frames at dense\,FPS. Progressive expansion to each subsequent window is conditioned on the preceding window's mean score exceeding $\theta_{\text{trigger}}$.
For tier-aware fusion (Eq.~\ref{eq:lambda-tiers}), we set $\lambda_{\text{base}} = 0.3$, $\lambda_{\text{motion}} = 0.4$, and $\lambda_{\text{anchor}} = 0.7$ across all datasets. These values reflect a monotonically increasing trust in the frame-level stream as the sampling tier rises.
We smooth $\lambda_t$ with a 1-D Gaussian ($\sigma_{\lambda} = 0.25$\,s) and clip to $[0,1]$ before applying Eq.~\ref{eq:fusion} to avoid discontinuities at tier boundaries. The fused score $\tilde{s}_t$ is then post-processed with a second 1-D Gaussian filter ($\sigma = 1.0$\,s) for evaluation stability.

\begin{table*}[!t]
\centering
\captionsetup{skip=4pt,labelfont={color=captionblue},textfont={color=captionblue}}
\caption{VAD Performance comparison across XD-Violence, UCF-Crime, MSAD and ShanghaiTech. \textcolor{green!70!black}{\ding{51}} / \textcolor{red}{\ding{55}} indicate whether a method is zero-shot and training-free. {\footnotesize Our MoG-VLM results are mean $\pm$ std over 3 runs.}}
\vspace{2pt}
\label{tab:vad_comparison}
\small
\setlength{\tabcolsep}{3.5pt}
\renewcommand{\arraystretch}{1.05}
\resizebox{\textwidth}{!}{%
\begin{tabular}{lccccccc}
\toprule
\multirow{2}{*}{Method} & \multirow{2}{*}{Zero-shot} & \multirow{2}{*}{Training-free} & XD-Violence & UCF-Crime & MSAD & ShanghaiTech \\
\cmidrule(lr){4-4} \cmidrule(lr){5-5} \cmidrule(lr){6-6} \cmidrule(lr){7-7}
 & & & AUC(\%) & AUC(\%) & AUC(\%) & AUC(\%) \\
\midrule

FusedVision~\cite{dawoud2025fusedvision} & \textcolor{red}{\ding{55}} & \textcolor{red}{\ding{55}} & -- & -- & -- & 83.58 \\
RTFM~\cite{tian2021weakly} & \textcolor{red}{\ding{55}} & \textcolor{red}{\ding{55}} & 61.56 & -- & \textbf{85.67} & -- \\
Jigsaw~\cite{wang2022video} & \textcolor{red}{\ding{55}} & \textcolor{red}{\ding{55}} & -- & -- & -- & 84.30 \\
MGFN~\cite{chen2023mgfn} & \textcolor{red}{\ding{55}} & \textcolor{red}{\ding{55}} & -- & 86.98 & \underline{85.0} & -- \\
STPrompt~\cite{wu2024weakly} & \textcolor{red}{\ding{55}} & \textcolor{red}{\ding{55}} & -- & 88.08 & -- & -- \\
OVVAD~\cite{wu2024open} & \textcolor{red}{\ding{55}} & \textcolor{red}{\ding{55}} & -- & 86.40 & -- & -- \\
Holmes-VAU~\cite{zhang2025holmes} & \textcolor{red}{\ding{55}} & \textcolor{red}{\ding{55}} & -- & 88.96 & -- & -- \\
MULDE~\cite{micorek2024mulde} & \textcolor{red}{\ding{55}} & \textcolor{red}{\ding{55}} & -- & - & -- & \underline{86.70} \\
LEC-VAD ~\cite{wang2025learning} & \textcolor{red}{\ding{55}} & \textcolor{red}{\ding{55}} & 88.47 & \underline{89.97} & -- & -- \\
Yu Wang \etal ~\cite{Learning_Local_Semantic} & \textcolor{red}{\ding{55}} & \textcolor{red}{\ding{55}} & \underline{88.53} & 89.75 & -- & \textbf{98.13} \\
LAS-VAD~\cite{Wang_2026_CVPR} & \textcolor{red}{\ding{55}} & \textcolor{red}{\ding{55}} & \textbf{89.96} & \textbf{91.05} & -- & -- \\
AnomalyRuler~\cite{yang2024follow} & \textcolor{red}{\ding{55}} & \textcolor{green!70!black}{\ding{51}} & -- & -- & -- & 85.20 \\
VERA~\cite{ye2025vera} & \textcolor{red}{\ding{55}} & \textcolor{green!70!black}{\ding{51}} & 88.26 & 86.55 & -- & -- \\
\midrule

LLAVA-1.5~\cite{liu2024improved} (ZS) & \textcolor{green!70!black}{\ding{51}} & \textcolor{green!70!black}{\ding{51}} & 79.62 & 72.84 & -- & -- \\
UR-DMU~\cite{zhou2023dual} & \textcolor{green!70!black}{\ding{51}} & \textcolor{green!70!black}{\ding{51}} & -- & -- & 74.30 & -- \\
Eventvad~\cite{shao2025eventvad} & \textcolor{green!70!black}{\ding{51}} & \textcolor{green!70!black}{\ding{51}} & 87.51 & 82.30 & -- & -- \\
AnyAnomaly~\cite{ahn2026anyanomaly} & \textcolor{green!70!black}{\ding{51}} & \textcolor{green!70!black}{\ding{51}} & -- & 80.70 & -- & \textbf{79.70} \\
LAVAD~\cite{zanella2024harnessing} & \textcolor{green!70!black}{\ding{51}} & \textcolor{green!70!black}{\ding{51}} & 85.36 & 80.28 & -- & -- \\
Unified-reasoning~\cite{lin2025unified} & \textcolor{green!70!black}{\ding{51}} & \textcolor{green!70!black}{\ding{51}} & \underline{91.34} & \underline{84.28} & \underline{85.90} & -- \\
\midrule
MoG-VLM (adaptive) & \textcolor{green!70!black}{\ding{51}} & \textcolor{green!70!black}{\ding{51}} & \textbf{91.91 ± .10 } & \textbf{84.30 ± .08} & \textbf{87.88 ± .17 } & \underline{78.96 ± .14} \\

\bottomrule
\end{tabular}%
}
\end{table*}

\subsection{Results}
\subsubsection{Comparison to State-of-the-Art Methods and Discussion}
\label{sec:sota_results}
Table~\ref{tab:vad_comparison} compares MoG-VLM against representative weakly-supervised, 
one-class, and zero-shot training-free baselines on four standard benchmarks. 
Among methods that are simultaneously zero-shot and training-free the regime in 
which our method operates, MoG-VLM establishes new state of the art on three of 
the four benchmarks. On XD-Violence, we reach $91.91\%$ AUC-ROC, improving over 
the previous best zero-shot training-free method, Unified-reasoning~\cite{lin2025unified}, 
by $+0.57\%$ while simultaneously delivering a $1.31\times$ end-to-end inference 
speedup (Sec.~\ref{sec:efficiency}). On MSAD, 
we obtain $87.88\%$, a 
$+1.98\%$ absolute gain over Unified-reasoning ($85.90\%$) and a $+2.88\%$ gain 
over the strongest non-zero-shot method, MGFN~\cite{chen2023mgfn}. 
On UCF-Crime, MoG-VLM reaches $84.30\%$, marginally surpassing Unified-reasoning
($84.28\%$). It also exceeds other zero-shot, training-free baselines such as
LAVAD ($80.28\%$), AnyAnomaly ($80.70\%$), and EventVAD ($82.30\%$)
by a clear margin.

These gains are particularly notable given the constraints of our setting. 
Methods such as Holmes-VAU~\cite{zhang2025holmes} ($88.96\%$ on UCF-Crime) and 
STPrompt~\cite{wu2024weakly} ($88.08\%$) rely on weak supervision with 
video-level labels and dataset-specific tuning, while VERA~\cite{ye2025vera} 
performs verbalized learning that adapts prompts to each benchmark. 
On XD-Violence, 
our zero-shot training-free result of $91.91\%$ in fact exceeds all supervised 
baselines we are aware of, suggesting that motion-adaptive sampling and 
anchor-triggered refinement provide a complementary inductive bias that learned 
representations alone do not capture.

Fig.\ref{fig:qualitative_breakdown} disaggregates detection performance by anomaly category. On MSAD and XD, most categories exceed 90\% AUC-ROC, with near-perfect scores on high-motion events such as Fighting and Abuse. UCF exhibits greater variance, visually distinctive events like Assault (94.8\%) are detected reliably, while categories involving minimal distinguishing motion (e.g., Shoplifting, 59.8\%) remain challenging, as the anomalous action is often brief or subtle. The results suggest that detection is most reliable when anomalies manifest as clearly  dynamic events. Shang is omitted as its abnormal events are not annotated by category.

\begin{table*}[t]
\centering
\footnotesize
\setlength{\tabcolsep}{3pt}
\renewcommand{\arraystretch}{1.10}
\captionsetup{skip=4pt}

\begin{minipage}[t]{0.62\textwidth}
\centering
\captionsetup{skip=4pt,labelfont={color=captionblue},textfont={color=captionblue}}
\caption{Breakdown of AUC-ROC (\%) across datasets under different components.}
\label{tab:breakdown_ablation}
\begin{tabularx}{\textwidth}{l
    >{\columncolor{framecol}\centering\arraybackslash}X
    >{\columncolor{framecol}\centering\arraybackslash}X
    >{\columncolor{framecol}\centering\arraybackslash}X
    >{\columncolor{temporalcol}\centering\arraybackslash}X
    >{\columncolor{dualcol}\centering\arraybackslash}X
    >{\columncolor{refinecol}\centering\arraybackslash}X}
\toprule
Dataset &
\shortstack{Uniform\\baseline\\$r_{\text{base}}$\\(Eq.~\ref{eq:TQM})} &
\shortstack{+\\(TGM)\\(Eq.~\ref{eq:TQM})} &
\shortstack{+Frame\\anchor\\(Eq.~\ref{eq:hierarchical_scoring})} &
\shortstack{Temporal\\only\\(Eq.~\ref{eq:temporal_scoring})} &
\shortstack{Fusion\\(Eq.~\ref{eq:fusion})} &
\shortstack{Refine-\\ment} \\
\midrule
XD    & 84.00 & 84.69 & 87.29 & 85.17 & 88.60 & \textbf{91.91} \\
UCF   & 65.02 & 68.31 & 70.07 & 74.48 & 77.99 & \textbf{84.30} \\
MSAD  & 78.41 & 81.51 & 82.31 & 79.06 & 84.84 & \textbf{87.88} \\
Shang & 71.54 & 72.38 & 73.10 & 61.64 & 75.87 & \textbf{78.96} \\
\bottomrule
\end{tabularx}
\end{minipage}
\hfill
\begin{minipage}[t]{0.36\textwidth}
\centering
\captionsetup{skip=4pt,labelfont={color=captionblue},textfont={color=captionblue}}
\caption{Per-component computational cost.
}
\label{tab:computational}
\begin{tabular}{@{}lcc@{}}
\toprule
\textbf{Component} & \textbf{Bat} & \textbf{s/b} \\
\midrule
\multicolumn{3}{@{}l}{\textit{Stream\,A --- Frame-Level}} \\
\;\;VLM Cap.\ (Qwen3VL-8B) & 24 & 8.19 \\
\;\;LLM Score (Qwen3-4B)     & 24 & 0.64 \\
\midrule
\multicolumn{3}{@{}l}{\textit{Stream\,B --- Temporal-Level}} \\
\;\;Vid-VLM (VideoLLaMA3)    &  4 & 2.04 \\
\;\;LLM Score (Qwen3-4B)     & 24 & 0.64 \\
\bottomrule
\end{tabular}
\end{minipage}

\end{table*}

\subsubsection{Progressive Pipeline Component Analysis}
\label{sec:component_analysis}

Table~\ref{tab:breakdown_ablation} decomposes the contribution of each component by 
progressively activating them on top of a uniform-sampling baseline, tracing 
the full path to the final MoG-VLM.

\noindent\textbf{Motion-adaptive sampling (TGM).} The TGM gate improves AUC-ROC by $+0.69\%$ to $+3.29\%$ across all datasets relative to the uniform baseline. These consistent gains confirm that motion gating successfully concentrates VLM computation on informative intervals without sacrificing detection quality.
\textbf{Anchor-triggered hierarchical refinement.} Adding anchor refinement produces strongly dataset-dependent effects. On UCF ($+1.76\%$), XD ($+2.6\%$) and MSAD ($+0.80\%$), hierarchical lookback contributes substantially, as many anomalies in these benchmarks (assault, accidents) unfold over multi-second pre-event contexts that benefit from denser temporal evidence. 
\textbf{Temporal stream.}
The temporal column reveals notably different behavior across datasets. On UCF, the video stream alone ($74.48\%$) surpasses the frame stream ($70.07\%$) by $+4.41\%$, suggesting that action-level dynamics carry strong discriminative signal for the diverse activity categories in this benchmark. On Shang, in contrast, the temporal stream ($61.64\%$) trails the uniform baseline ($71.54\%$), consistent with a domain gap between VideoLLaMA3's pretraining distribution and the subtle pedestrian-scene anomalies (skating, wrong-direction walking) that characterize this dataset. Tier-aware fusion is designed to accommodate precisely this kind of cross-dataset variation, allowing each stream to contribute where it is most reliable.
\textbf{Tier-aware dual-stream fusion.} 
Fusing the two streams via the tier-conditioned coefficient $\lambda_t$ (Eq.~\ref{eq:fusion})  yields the largest single improvement at any stage on two of four datasets: +7.92\% on UCF, and +2.77\% on Shang, with a smaller +1.31\% on XD where anchor refinement contributes more.
The same mechanism, with 
identical scalars $(\lambda_{\text{base}}, \lambda_{\text{motion}}, 
\lambda_{\text{anchor}})$, accommodates both directions of complementarity. On 
UCF the temporal stream rescues the frame stream, while on Shang 
the frame stream rescues a temporal stream that performed below baseline. 
Because $\lambda_t$ inherits from the motion sampling tier rather than being 
tuned per dataset, anchor-tier frames automatically receive the largest 
frame-stream weight while base-tier frames defer to the temporal stream. \textbf{Score refinement.} Following standard VAD post-processing 
practice~\cite{georgescu2021anomaly, lin2025unified, ye2025vera}, the 
fused score $\tilde{s}_t$ and the tier coefficient $\lambda_t$ are 
passed through a 1D Gaussian filter to suppress high-frequency noise 
from VLM caption variability and tier transitions. This step introduces 
no new evidence and is applied identically by our zero-shot baselines, but is required for fair comparison.
\begin{table*}[t]
\centering
\footnotesize

\begin{minipage}[t]{0.48\textwidth}
\vspace{0pt}\centering
\captionsetup{skip=4pt,labelfont={color=captionblue},textfont={color=captionblue}}
\captionof{table}{Cumulative timing breakdown (hours) across datasets.}
\label{tab:stage_breakdown}
\setlength{\tabcolsep}{3pt}
\renewcommand{\arraystretch}{1.10}
\begin{tabularx}{\linewidth}{@{}l*{4}{>{\centering\arraybackslash}X}@{}}
\toprule
Configuration & XD & UCF & MSAD & Shang\\
\midrule
Uniform (1 FPS)            & 7.96 & 3.51 & 0.78 & 0.13 \\
\,+ (TGM)                  & 8.42 & 4.12 & 0.85 & 0.14 \\
\,+ Frame anchor           & 9.60 & 5.21 & 0.96 & 0.17 \\
Temporal only              & 5.20 & 2.59 & 0.66 & 0.11 \\
\midrule
Final                       & \textbf{14.80} & \textbf{7.80} & \textbf{1.62} & \textbf{0.28} \\
\bottomrule
\end{tabularx}

\end{minipage}%
\hfill
\begin{minipage}[t]{0.48\textwidth}
\vspace{0pt}\centering
\captionsetup{skip=4pt,labelfont={color=captionblue},textfont={color=captionblue}}
\captionof{table}{Throughput across datasets.}
\label{tab:throughput}
\scriptsize
\setlength{\tabcolsep}{1pt}
\renewcommand{\arraystretch}{1.00}
\setlength{\aboverulesep}{0pt}
\setlength{\belowrulesep}{0pt}
\setlength{\abovetopsep}{0pt}
\setlength{\belowbottomsep}{0pt}
\begin{tabular*}{\linewidth}{@{\extracolsep{\fill}}llrrr@{}}
\toprule
\textbf{Dataset} & \textbf{Method} & \textbf{Time (h)} & \textbf{FPS} & \textbf{Speedup} \\
\midrule
\multirow{3}{*}{XD}
 & LAVAD~\cite{zanella2024harnessing}   & 71.40 & 9.1   & 1.0$\times$ \\
 & Unified~\cite{lin2025unified}        & 19.40 & 33.4  & 3.7$\times$ \\
 & \textbf{Ours}                        & \textbf{14.80} & \textbf{43.84} & \textbf{4.82$\times$} \\
\midrule
\multirow{3}{*}{UCF}
 & LAVAD~\cite{zanella2024harnessing}   & 33.93 & 9.1   & 1.0$\times$ \\
 & Unified~\cite{lin2025unified}        &  9.24 & 33.40 & 3.7$\times$ \\
 & \textbf{Ours}                        & \textbf{7.80} & \textbf{39.25} & \textbf{4.35$\times$} \\
\midrule
\multirow{3}{*}{MSAD}
 & LAVAD~\cite{zanella2024harnessing}   & 6.95 & 9.1   & 1.0$\times$ \\
 & Unified~\cite{lin2025unified}        & 1.90 & 33.40 & 3.7$\times$ \\
 & \textbf{Ours}                        & \textbf{1.62} & \textbf{39.01} & \textbf{4.30$\times$} \\
\midrule
\multirow{3}{*}{Shang}
 & LAVAD~\cite{zanella2024harnessing}   & 1.24 & 9.1   & 1.0$\times$ \\
 & Unified~\cite{lin2025unified}        & 0.34 & 33.40 & 3.7$\times$ \\
 & \textbf{Ours}     & \textbf{0.28} & \textbf{40.46} & \textbf{4.45$\times$} \\
\bottomrule
\end{tabular*}
\end{minipage}
\end{table*}
%
\subsection{Qualitative Results}
\begin{figure*}[t]
\centering
\captionsetup{skip=4pt,labelfont={color=captionblue},textfont={color=captionblue}}

\begin{minipage}{\textwidth}
\centering
\includegraphics[width=\textwidth]{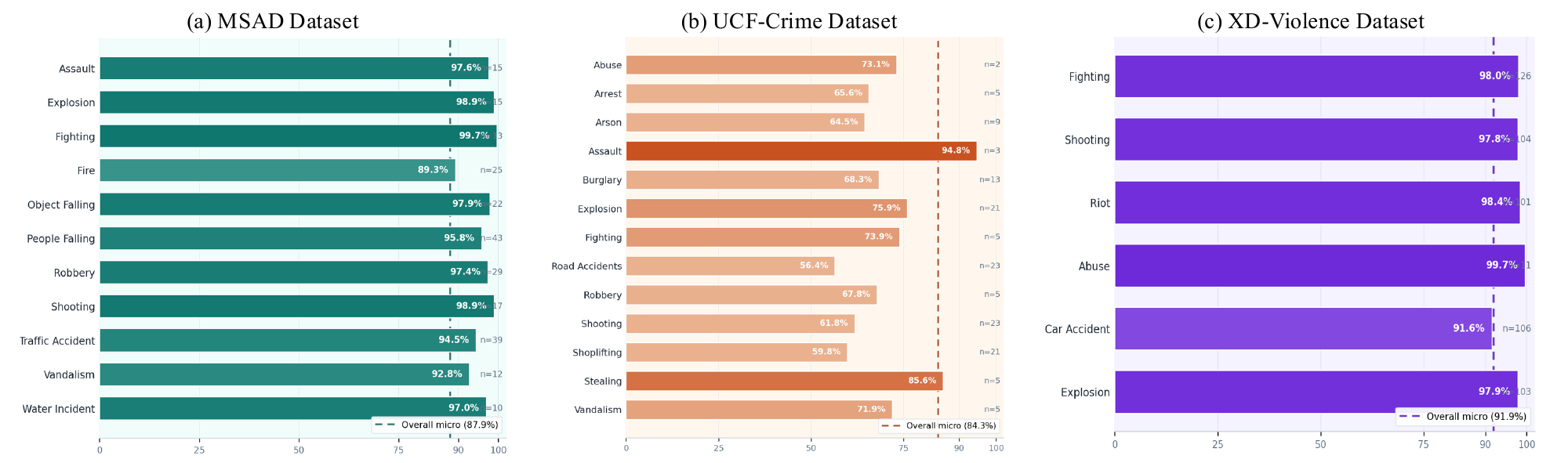}
\captionof{figure}{Per-category AUC-ROC breakdown across three benchmarks. 
}
\label{fig:qualitative_breakdown}
\end{minipage}

\vspace{4pt}

\begin{minipage}{\textwidth}
\centering
\includegraphics[width=\textwidth]{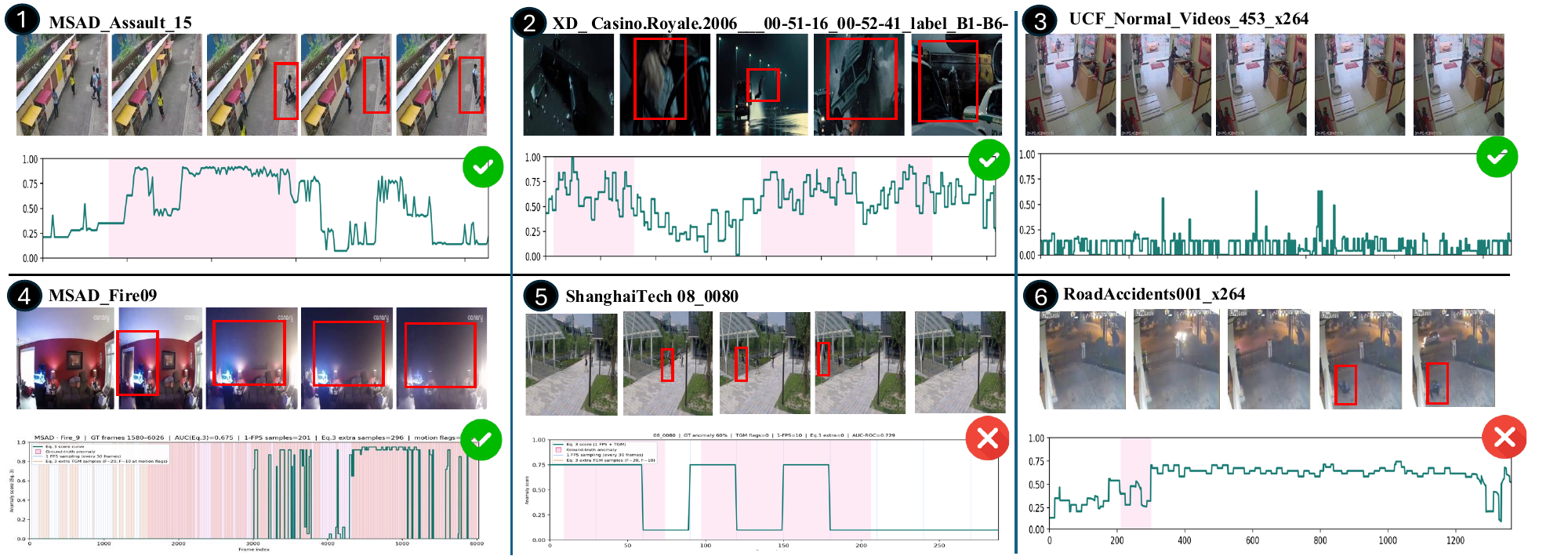}
\captionof{figure}{Qualitative results. Pink regions denote ground-truth anomalous intervals.}
\label{fig:Qualitative}
\end{minipage}
\end{figure*}

Figure~\ref{fig:Qualitative} presents per-frame anomaly score trajectories for six videos spanning distinct anomaly regimes.
\textbf{MSAD---Assault (1).} Both streams correctly elevate scores during the annotated anomaly window. Notably, the fused output sustains elevated scores beyond the ground-truth boundary, capturing the post-impact interval in which the victim remains prone on the ground after frame~\#150, a behaviorally meaningful detection that extends beyond the ground truth.
\textbf{XD-Violence---Multi-category violence (2).} Across six concatenated anomaly types interspersed with normal intervals, both streams suppress their outputs during the normal segment (frames 300--1000) and re-elevate in close temporal alignment with each subsequent event, confirming that tier-aware fusion generalizes across heterogeneous anomaly semantics without category-specific tuning. \textbf{UCF--Normal Videos-453-264 (3)} The  score remains consistently near zero throughout, demonstrating stable false-positive suppression under normal conditions. \textbf{MSAD---Fire09 (4)} TGM gating remains robust to photometric perturbations: a flickering television and ambient illumination shifts in the normal interval stay below the z-score threshold, while the propagating flame and smoke sustain a coherent multiscale gradient signal that recruits additional VLM inference, improving AUC-ROC from 0.648 (1 FPS) to 0.675 (Sec.~\ref{sec:adaptive_sampling}). \textbf{ShanghaiTech---08\_0080 (5)} The anomalous skating behavior blends visually with surrounding pedestrian activity, and TGM fails to intensify on a motion pattern that is kinematically indistinguishable from normal walking, with AUC-ROC plateauing at the 1 FPS baseline of 72.9\% and showing that brief, subtle ShanghaiTech anomalies remain challenging. \textbf{UCF---RoadAccidents001 (6).} The brief hit is missed, but the model sustains an elevated score throughout the prolonged post-impact interval in which the victim remains prone, indicating that sudden short-lived anomalies can evade detection while their behavioral post-event manifestation is more reliably captured.

\subsection{Computational Efficiency Analysis}
\label{sec:efficiency}
We report timing across XD, UCF, MSAD and Shang on a single NVIDIA A100 40\,GB GPU. Table~\ref{tab:computational} gives per-component latency, Table~\ref{tab:stage_breakdown} traces cumulative timing cost as each stage is activated, and Table~\ref{tab:throughput} compares end-to-end throughput against prior zero-shot training-free methods.
\textbf{Where time is spent.} Frame-level VLM captioning dominates at 8.19\,s/batch (Qwen3-VL-8B), followed by VideoLLaMA3 at 2.04\,s/batch; LLM scoring is negligible (0.64\,s/batch). Pipeline cost is therefore governed by how often the  visual encoders are invoked.
\textbf{Cost attribution per stage.} The 1\,FPS uniform baseline consumes 7.96, 3.51, 0.78 and .13\,h on XD, UCF, MSAD and Shang. TGM gating adds 6--17\%, anchor refinement adds a further 13--26\%, and the temporal stream contributes roughly half the frame-stream cost. The final dual configuration totals 14.80, 7.80, 1.62 and .28\,h.
\textbf{Throughput.} Following the convention of stride-based methods~\cite{lin2025unified}, we report amortized effective throughput $\mathrm{FPS}_{\mathrm{eff}} = N_{\mathrm{frames}}/T_{\mathrm{pipeline}}$. MoG-VLM reaches 43.84, 36.25, 39.01 and 40.46\,FPS, yielding $4.30$--$4.82\times$ speedups over LAVAD and $1.17$--$1.31\times$ over Unified-reasoning. On XD-Violence, it saves 4.60\,h of GPU inference time while improving AUC-ROC by $+0.57\%$.
\textbf{Where the savings originate.} Three mechanisms: a coarser 1\,FPS baseline stride (vs. stride-16 in prior work) nearly halves the candidate pool before any content-aware filtering; TGM gating suppresses VLM inference on static intervals; and anchor expansion triggers denser lookback only when frame-level scores exceed $\theta_{\mathrm{trigger}}$, with early termination.

Crucially, motion-adaptive allocation improves both axes of the efficiency and accuracy trade-off jointly: a $1.17$--$1.31\times$ speedup over Unified-reasoning is achieved alongside AUC-ROC gains on every benchmark where direct comparison is reported, indicating that the savings originate from \emph{where} compute is spent rather than \emph{how much}. The resulting $36$--$44$\,FPS throughput on a single A100 enters the operating regime of standard surveillance camera framerates, suggesting that content-aware allocation is a more deployable axis of scaling for VLM-based VAD than uniformly densifying inference.

\subsection{Ablation}
\noindent\textbf{Multiscale Window Size Sensitivity.}
We ablate the temporal window sizes used for multiscale TGM computation by comparing three configurations: $\mathcal{W} = \{1,2,4\}$, $\{1,3,5\}$, and $\{2,4,8\}$. Our default  $\mathcal{W} = \{1,2,4\}$ was fixed before this sensitivity study. Table~\ref{tab:window_ablation} reports frame-level AUC-ROC on all benchmarks (TGM only). Performance is stable across configurations, with the largest gap being 2.48\% on MSAD. The wider $\mathcal{W}=\{2,4,8\}$ performs best on UCF and XD, while MSAD and Shang favors $\mathcal{W}=\{1,2,4\}$, suggesting that aggregating motion over longer horizons benefits scene-diverse benchmarks while finer granularity captures short pedestrian-scene events. The robustness across configurations indicates that the framework is not sensitive to this hyperparameter choice.
\textbf{Anchor Trigger Threshold Sensitivity.} Table~\ref{tab:theta_trigger} shows that performance peaks at lower thresholds ($\theta_{\text{trigger}} = 0.4$--$0.5$) across all datasets, with AUC-ROC degrading as $\theta_{\text{trigger}}$ increases beyond $0.5$. Most notably a drop of $2.76\%$ on XD and $2.60\%$ on UCF when raising the threshold from $0.4$ to $0.8$. This indicates that permissive anchor activation is preferable, as overly conservative thresholds suppress hierarchical refinement. See Supplementary for more ablation studies. 
\textbf{We include ablations} on tier-aware fusion weights, EMA freezing during anomaly windows, $\theta_{\text{anom}}$ sensitivity, and Gaussian smoothing strength in Supplementary Section~A.
\begin{figure*}[t]
\centering
\captionsetup{skip=4pt,labelfont={color=captionblue},textfont={color=captionblue}}
\begin{tabular}{@{}p{0.49\textwidth}@{\hspace{0.02\textwidth}}p{0.47\textwidth}@{}}
\begin{minipage}[t]{\linewidth}
\vspace{0pt}\centering\footnotesize
\captionof{table}{Ablation of multiscale TGM window sizes $\mathcal{W}$. AUC-ROC (\%) is reported at the frame-level (motion-adaptive sampling only). The default configuration is shaded.}
\label{tab:window_ablation}
\setlength{\tabcolsep}{6pt}
\renewcommand{\arraystretch}{1.10}
\begin{tabularx}{\linewidth}{@{}l*{4}{>{\centering\arraybackslash}X}@{}}
\toprule
$\mathcal{W}$ & XD & UCF & MSAD & Shang \\
\midrule
\rowcolor{gray!15}
$\{1, 2, 4\}$ & 84.69 & 68.31 & \textbf{81.51} & \textbf{72.38} \\
\rowcolor{white}
$\{1, 3, 5\}$ & 84.53 & 66.76 & 79.07 & 71.38 \\
$\{2, 4, 8\}$ & \textbf{85.36} & \textbf{68.61} & 79.38 & 70.67 \\
\bottomrule
\end{tabularx}
\end{minipage}
&
\begin{minipage}[t]{\linewidth}
\vspace{0pt}\centering\footnotesize
\captionof{table}{Effect of $\theta_{\text{trigger}}$ on AUC-ROC (\%) across datasets. Default conf. is shaded.}
\label{tab:theta_trigger}
\setlength{\tabcolsep}{3pt}
\renewcommand{\arraystretch}{1.10}
\begin{tabularx}{\linewidth}{@{}c*{4}{>{\centering\arraybackslash}X}@{}}
\toprule
$\theta_{\text{trigger}}$ & XD & UCF & MSAD & Shang \\
\midrule
0.4 & \textbf{87.56} & \textbf{70.21} & \textbf{82.92} & \textbf{73.44} \\
\rowcolor{gray!15}
0.5 & 87.29 & 70.07 & 82.31 & 73.10 \\
\rowcolor{white}
0.6 & 85.84 & 68.46 & 82.22 & 73.09 \\
0.7 & 85.29 & 67.95 & 82.14 & 73.16 \\
0.8 & 84.80 & 67.61 & 82.07 & 73.16 \\
\bottomrule
\end{tabularx}
\end{minipage}
\end{tabular}
\end{figure*}

\section{Conclusion}
We presented MoG-VLM, a zero-shot training-free framework that treats motion as a lightweight control signal governing when VLM inference is invoked. 
MoG-VLM sets state of the art on UCF-Crime, XD-Violence, and MSAD, remains competitive on ShanghaiTech, and achieves a speedup over the strongest zero-shot baseline, demonstrating that content-aware allocation of model capacity is a more productive axis of progress than scaling uniform VLM inference.

\section*{Acknowledgement}
This work is supported by MBZUAI Base funds BF0008 and by the BK21 FOUR Project (Bigdata Research and Education Group for Enhancing Social Connectedness Through Advanced Data Technology and Interaction Science Research, 5199990913845), funded by the Ministry of Education (MOE, Korea) and the National Research Foundation of Korea (NRF).

\bibliography{egbib}

\clearpage
\appendix

\title{Supplementary Material for MoG-VLM: Efficient Video Anomaly Detection through Motion-Guided Vision-Language Models}
\runninghead{Dawoud, ET AL}{MoG-VLM}
\addauthor{Khaled Waleed Dawoud
}{khaled.dawoud@mbzuai.ac.ae}{1}
\addauthor{Muhammad Zaigham Zaheer}{Zaigham.Zaheer@mbzuai.ac.ae}{1}
\addauthor{Khan Muhammad}{khanmuhammad@skku.edu}{2}
\addauthor{Muhammad Haris Khan}{muhammad.haris@mbzuai.ac.ae}{1}

\addinstitution{
 Mohamed bin Zayed University of Artificial Intelligence,\\
 Abu Dhabi,\\
 United Arab Emirates
}
\addinstitution{
Sungkyunkwan University,\\
Seoul 03063,\\
South Korea
}

\maketitle

\section*{Appendix Roadmap}
In this appendix, we cover the following materials:
\begin{description}[leftmargin=2.5em, labelwidth=2em]
  \item[A] Extended ablations
  \begin{description}[leftmargin=2.5em, labelwidth=2em]
    \item[A.1] Tier-aware fusion weight sensitivity \dotfill S1
    \item[A.2] EMA freezing during anomaly windows \dotfill S2
    \item[A.3] Anomaly threshold $\theta_{\text{anom}}$ sensitivity \dotfill S2
    \item[A.4] Gaussian smoothing strength \dotfill S2
    \item[A.5] Prompt sensitivity \dotfill S2
    \item[A.6] Isolating the sampling contribution \dotfill S2
  \end{description}
  \item[B] Additional qualitative results \dotfill S3
  \item[C] Prompts \dotfill S4
  \begin{description}[leftmargin=2.5em, labelwidth=2em]
    \item[C.1] Frame-level captioning prompt \dotfill S4
    \item[C.2] Frame-level anomaly scoring prompt \dotfill S5
    \item[C.3] Temporal multi-frame captioning prompt \dotfill S5
  \end{description}
  \item[D] Limitations \dotfill S6
\end{description}

\section{Extended Ablations}

\subsection{Tier-Aware Fusion Weight Sensitivity}
\label{sec:ablation_lambda}

We sweep each tier weight in $(\lambda_{\text{base}}, \lambda_{\text{motion}}, \lambda_{\text{anchor}})$ independently while holding the other two at the paper configuration $(0.3, 0.4, 0.7)$. Results are reported in Tables~\ref{tab:lambda_anchor}--\ref{tab:lambda_motion}.

\textbf{$\lambda_{\text{anchor}}$} dominates the trade-off. Raising it from 0.2 to 0.7 lifts mean AUC by 0.63\% driven by a 2.31\% gain on ShanghaiTech against a 0.28\% drop on UCF-Crime, XD-Violence is invariant and MSAD gains 0.50\%. Anchor-triggered frames thus carry stronger localization signal on sparse-anomaly scenes, but over-commit on the denser, more ambiguous UCF-Crime distribution.
\textbf{$\lambda_{\text{base}}$} exhibits the same asymmetry more sharply. Increasing it from 0.3 to 0.6 raises mean AUC to 86.31\% (+0.46\%), with ShanghaiTech gaining 1.54\% and UCF-Crime losing 0.55\%. We retain $\lambda_{\text{base}}=0.3$ because higher values inflate the mean at UCF-Crime's expense, where motion-gated triggering is the primary contributor.

\textbf{$\lambda_{\text{motion}}$ }is the least sensitive: the full sweep spans only 0.19\% in mean AUC, with per-dataset variation under 1.2\%. We keep $\lambda_{\text{motion}}=0.4$ as the inflection point where ShanghaiTech and UCF-Crime trends cross.

\begin{table}[h]
\centering
\footnotesize
\setlength{\tabcolsep}{6pt}
\renewcommand{\arraystretch}{1.1}
\begin{minipage}[t]{0.49\linewidth}
\centering
\captionsetup{font=scriptsize, skip=4pt}
\caption{$\lambda_{\text{anchor}}$ sweep ($\lambda_{\text{base}}{=}0.3$, $\lambda_{\text{motion}}{=}0.4$).}
\label{tab:lambda_anchor}
\begin{tabular}{l|cccc}
\toprule
$\lambda_{\text{anchor}}$ & MSAD & UCF & XD & Shang. \\
\midrule
0.2 & 87.43 & \textbf{84.76} & 91.97 & 76.71 \\
0.3 & 87.59 & 84.71 & 91.97 & 77.58 \\
0.4 & 87.72 & 84.66 & 91.97 & 78.16 \\
0.5 & 87.82 & 84.60 & 91.97 & 78.58 \\
0.6 & 87.89 & 84.54 & 91.97 & 78.86 \\
0.7\,$\star$ & \textbf{87.93} & 84.48 & 91.97 & \textbf{79.02} \\
\bottomrule
\end{tabular}
\end{minipage}
\hfill
\begin{minipage}[t]{0.49\linewidth}
\centering
\captionsetup{font=scriptsize, skip=4pt}
\caption{$\lambda_{\text{motion}}$ sweep ($\lambda_{\text{base}}{=}0.3$, $\lambda_{\text{anchor}}{=}0.7$).}
\label{tab:lambda_motion}
\begin{tabular}{l|cccc}
\toprule
$\lambda_{\text{motion}}$ & MSAD & UCF & XD & Shang. \\
\midrule
0.2 & 87.92 & \textbf{84.61} & 91.96 & 78.25 \\
0.3 & \textbf{87.94} & 84.55 & 91.97 & 78.64 \\
0.4\,$\star$ & 87.93 & 84.48 & 91.97 & 79.02 \\
0.5 & 87.90 & 84.38 & 91.97 & 79.26 \\
0.6 & 87.85 & 84.23 & 91.97 & \textbf{79.42} \\
\bottomrule
\end{tabular}
\end{minipage}
\end{table}

\begin{table}[h]
\centering
\footnotesize
\setlength{\tabcolsep}{6pt}
\renewcommand{\arraystretch}{1.1}
\captionsetup{skip=6pt}
\caption{Sensitivity to $\lambda_{\text{base}}$ ($\lambda_{\text{motion}}{=}0.4$, $\lambda_{\text{anchor}}{=}0.7$). AUC (\%). $\star$: paper configuration.}
\label{tab:lambda_base}
\begin{tabular}{l|cccc}
\toprule
$\lambda_{\text{base}}$ & MSAD & UCF & XD & Shang. \\
\midrule
0.2 & 87.57 & 84.47 & 91.79 & 77.99 \\
0.3\,$\star$ & 87.93 & \textbf{84.48} & 91.97 & 79.02 \\
0.4 & 88.20 & 84.39 & 92.10 & 79.79 \\
0.5 & 88.40 & 84.20 & 92.17 & 80.35 \\
0.6 & \textbf{88.56} & 83.90 & \textbf{92.18} & \textbf{80.56} \\
\bottomrule
\end{tabular}
\end{table}

\subsection{Ablation: EMA Freezing during Anomaly Windows}
\label{supp:ema_freezing}

The adaptive z-score in manuscript Eq. 3 maintains running
statistics $\mu_t$ and $\sigma_t$ via EMA updates. Updating these at
every frame allows sustained anomalous motion to be absorbed into the
baseline: as the EMA tracks elevated TGM values during an ongoing
event, the z-score decays below $\theta_{\text{anom}}$ and suppresses
motion-gated densification mid-event. We prevent this self-suppression
by freezing $\mu_t$ and $\sigma_t$ whenever $z(t) > \theta_{\text{anom}}$,
so the reference distribution reflects only normal background dynamics.
Table~\ref{tab:freeze_ablation} reports frame-level AUC-ROC with and
without freezing. The mechanism yields consistent gains, largest on
MSAD (+2.01\,pp) and UCF-Crime (+1.95\,pp) where anomalies unfold over
multi-second intervals, and smaller on XD-Violence (+0.44\,pp) where
punctate events (gunshots, explosions) complete before the EMA can
adapt. The mechanism adds no inference cost and no hyperparameters.

\begin{table}[t]
\centering
\footnotesize
\setlength{\tabcolsep}{4pt}
\renewcommand{\arraystretch}{1.1}
\begin{minipage}[t]{0.48\linewidth}
\centering
\captionsetup{font=scriptsize, skip=4pt}
\caption{Effect of EMA freezing during anomaly windows on frame-level AUC-ROC (\%). Default config (Sec.~4.2).}
\label{tab:freeze_ablation}
\begin{tabular}{|l|>{\columncolor{gray!15}}c|c|c|}
\hline
Dataset & With freeze & No freeze & $\Delta$ (\%) \\
\hline
XD-Violence & \textbf{84.69} & 84.25 & +0.44 \\
\hline
UCF-Crime   & \textbf{68.31} & 66.36 & +1.95 \\
\hline
MSAD        & \textbf{81.51} & 79.50 & +2.01 \\
\hline
\end{tabular}
\end{minipage}
\hfill
\begin{minipage}[t]{0.48\linewidth}
\centering
\captionsetup{font=scriptsize, skip=4pt}
\caption{Sensitivity to $\theta_{\text{anom}}$ (Eq.~3). Frame-level AUC-ROC (\%). Default shaded.}
\label{tab:theta_anom_ablation}
\begin{tabular}{|l|c|c|c|c|}
\hline
$\theta_{\text{anom}}$ & XD. & UCF & MSAD & Shang. \\
\hline
1.5 & \textbf{85.64} & \textbf{68.41} & \textbf{81.68} & \textbf{72.50} \\
\hline
1.8 & 85.14 & 68.35 & 81.67 & 72.50 \\
\hline
2.0 & 85.01 & 68.26 & 81.61 & 72.50 \\
\hline
\rowcolor{gray!15}
2.3 (default) & 84.69 & 68.31 & 81.51 & 72.38 \\
\hline
\end{tabular}
\end{minipage}
\end{table}

\vspace{-.5cm}
\subsection{Ablation: Anomaly Threshold $\theta_{\text{anom}}$ Sensitivity}
\label{supp:theta_anom}
The motion gate in manuscript Eq.~3 triggers when the adaptive z-score exceeds $\theta_{\text{anom}}$. We fix
$\theta_{\text{anom}}=2.3$ across all datasets without per-dataset
tuning. \textbf{This value was selected prior to running the sensitivity analysis below and was not chosen to maximize performance on any dataset.} Table~\ref{tab:theta_anom_ablation} reports sensitivity over the range $[1.5, 2.3]$. Frame-level AUC-ROC varies by less than 1\% across this range on every benchmark, confirming that the framework is insensitive to this hyperparameter. All four datasets favor lower thresholds that admit more candidates.

\vspace{-.5cm}
\subsection{Ablation: Gaussian Smoothing Strength}
\label{sec:abl_smoothing}
\begin{figure}[t]
\centering
\includegraphics[width=.8\linewidth]{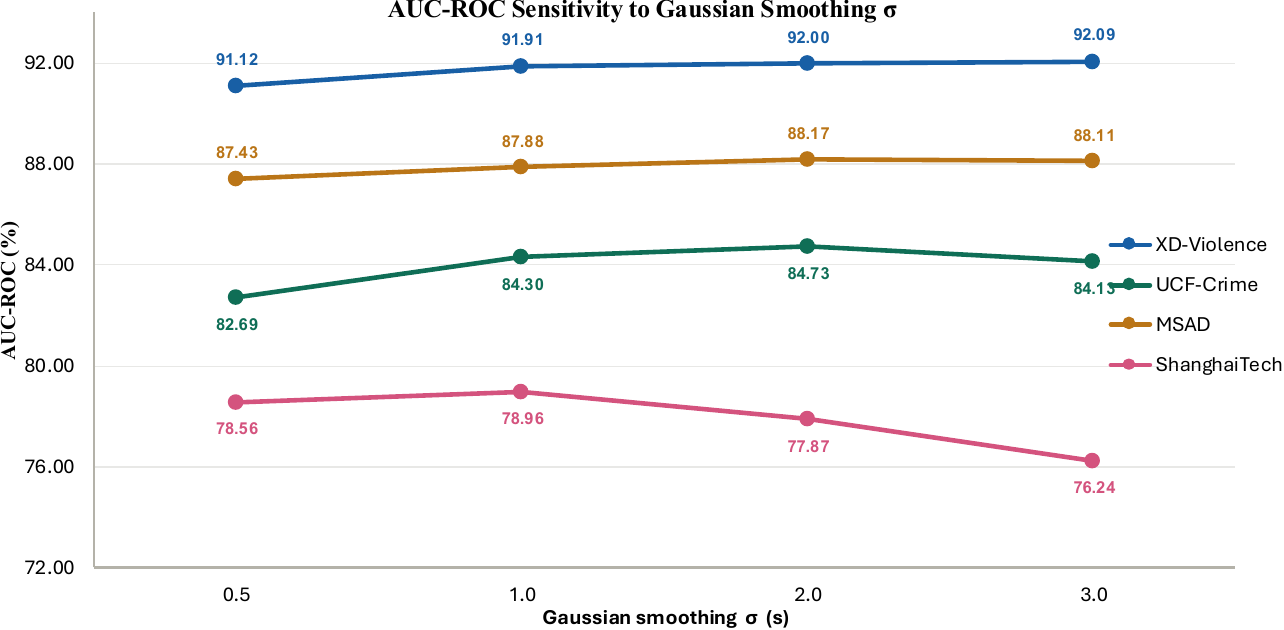}
\vspace{.5cm}
\caption{Sensitivity of frame-level AUC-ROC to the Gaussian smoothing width $\sigma$ across datasets.}
\label{fig:sigma_sweep}
\end{figure}

The fused score~$\tilde{s}_t$ is passed through a 1-D Gaussian filter to suppress high-frequency noise from caption variability and tier transitions. We sweep the filter width~$\sigma$ over $[0.5,\,3.0]$\,s with all other settings fixed; Fig.~\ref{fig:sigma_sweep} reports frame-level AUC-ROC. The optimal width is anomaly-density dependent. Sparse-anomaly ShanghaiTech degrades monotonically beyond $\sigma\!=\!1.0$, falling $2.32$\% to $76.24\%$ at $\sigma\!=\!3.0$ as brief events are blurred into the normal baseline. Dense-anomaly benchmarks are robust and mildly benefit from wider kernels: XD-Violence plateaus near $92.0\%$, MSAD peaks at $88.17\%$, and UCF-Crime at $84.73\%$, their multi-second intervals surviving while single-frame noise is removed. We fix $\sigma\!=\!1.0$ across all datasets, as it preserves the temporal precision sparse-anomaly detection requires.

\subsection{Ablation: Prompt Sensitivity}
The framework is insensitive to prompt wording. We evaluate four variants of the
frame-level pipeline prompts on XD dataset (Table~\ref{tab:R2_prompt_sensitivity}): the paper
default (full structured prompt), V1 (indicator lists removed), V2 (lists and rules cut, no structure), and V3 (default content reworded). AUC-ROC spans only
0.71\% across all four prompts, confirming that results do not depend on a hand-tuned
prompt. 

\begin{table}[h]
\centering
\setlength{\tabcolsep}{4pt}
\begin{tabular}{l c}
\toprule
Prompt variant & Final \\
\midrule
\rowcolor{gray!15}
Default (paper --- full prompt)             & 91.91 \\
V1 (minimal --- indicator lists removed)    & 91.98    \\
V2 (no structure --- lists + rules cut)     & 91.34 \\
V3 (rephrased --- same content, reworded)   & 92.05    \\
\bottomrule
\end{tabular}
\vspace{0.5em}
\caption{Prompt sensitivity on XD (AUC-ROC, \%).}
\label{tab:R2_prompt_sensitivity}
\end{table}

\subsection{Ablation: Isolating the Sampling Contribution}
To isolate motion-adaptive allocation, we hold the pipeline fixed and vary only the sampling schedule,
replacing it with uniform sampling at 3 FPS ($r_{\text{high}}$, MoG-VLM's densest
rate). Under this setting (Table~\ref{tab:R1_Samling}), MoG-VLM is
$2.55\text{--}2.86\times$ faster. Accuracy does not suffer:
frame-level AUC-ROC improves on XD-Violence, UCF-Crime, and MSAD, with a small
ShanghaiTech drop the full pipeline recovers. The gains thus stem from where compute
is spent.

\begin{table}[h]
\centering
\captionsetup{position=bottom}
\setlength{\tabcolsep}{4pt}
\begin{tabular}{lcccc}
\toprule
Sampling (AUC\%/h) & XD & UCF & MSAD & Shang \\
\midrule
Uniform 1~FPS & 84.00/7.96 & 65.02/3.51 & 77.57/0.78 & 71.54/0.13 \\
Uniform 3~FPS & 84.44/23.9 & 65.18/10.5 & 77.99/2.3 & \textbf{73.67}/0.4 \\
Ours (adaptive 1$\to$3 FPS) & \textbf{84.69}/8.42 & \textbf{68.31}/4.12 & \textbf{81.51}/0.85 & 72.38/0.14 \\
\bottomrule
\end{tabular}
\vspace{0.5em}
\caption{Sampling Schedule Ablation.}
\label{tab:R1_Samling}
\end{table}




\begin{figure*}[t]
\centering
\captionsetup{skip=6pt}

\includegraphics[width=\textwidth,height=0.3\textheight,keepaspectratio]{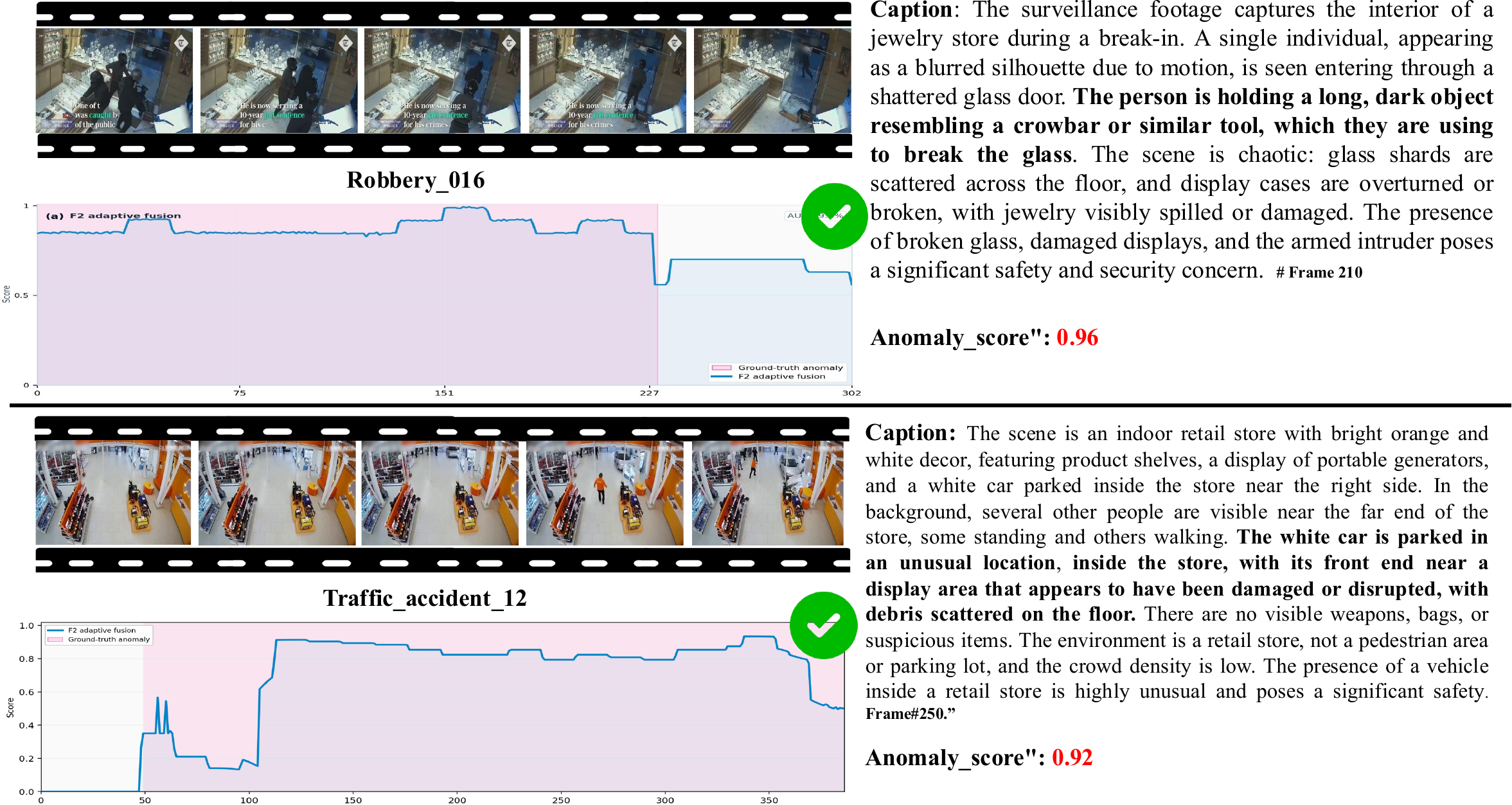}

\vspace{2mm}

\includegraphics[width=\textwidth,height=0.3\textheight,keepaspectratio]{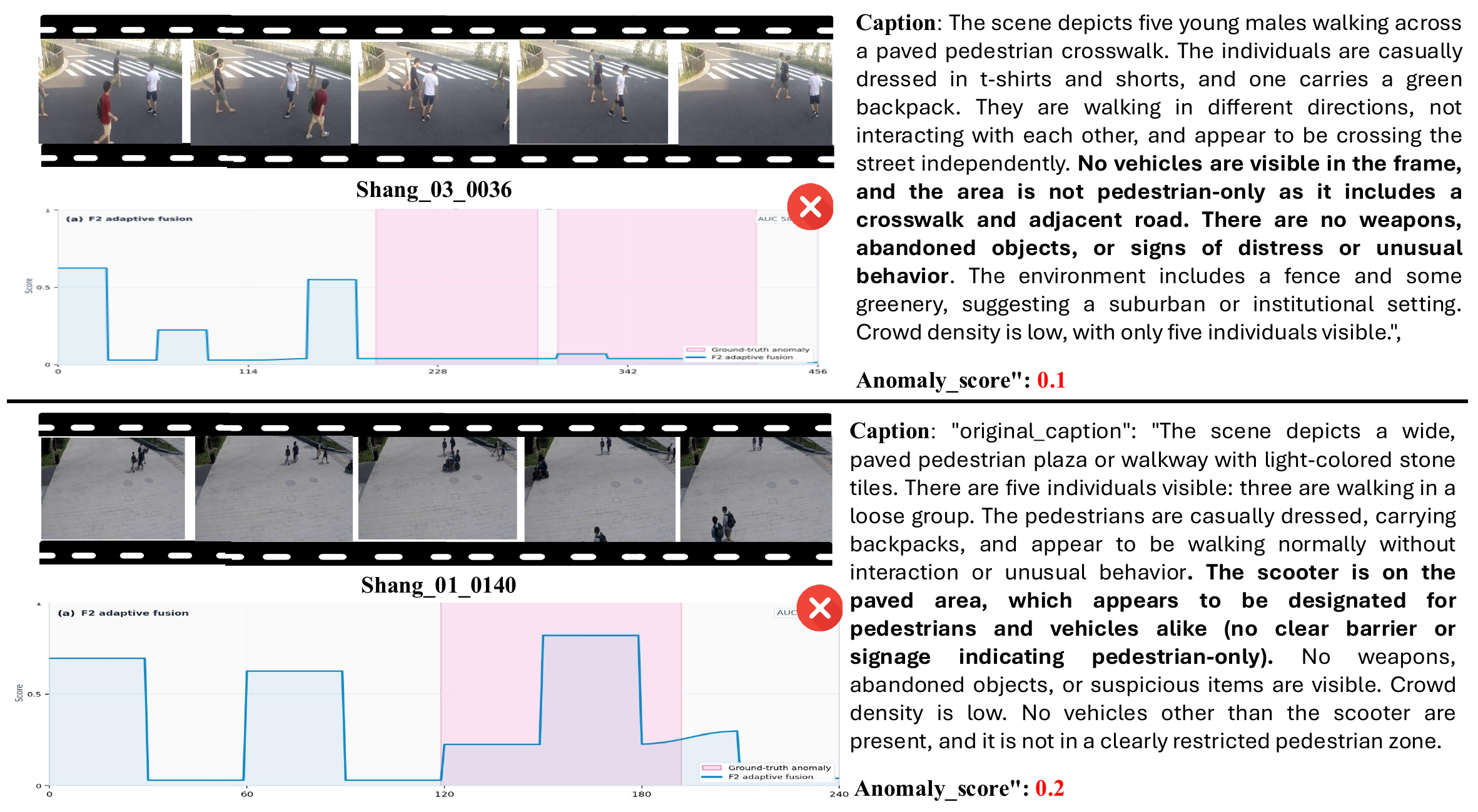}

\caption{Qualitative examples of successful detections and failure cases.}
\label{fig:qual_examples}
\end{figure*}

\section{Additional Qualitative Results}

\textbf{Successful Cases:} Fig.~\ref{fig:qual_examples} illustrates two successful detections on MSAD dataset. In Robbery\_016, forced entry through a glass door generates sustained high-magnitude motion that continuously identifies the weapon, structural damage, and intruder context, yielding a score of 0.96 that closely tracks the ground-truth anomaly boundary. Traffic\_accident\_12 presents a semantically anomalous event with comparatively modest motion, a vehicle driven into a retail interior. The score ramps gradually before stabilizing at a high-confidence plateau. Despite the absence of conventional violence cues, the scene-level caption resolves the anomaly through contextual incongruity, a vehicle inside a non-vehicular space, producing a score of
0.92. Together, these cases demonstrate MoG capacity to handle both kinematically and semantically driven anomalies.

\textbf{Failure Cases.} Fig.~\ref{fig:qual_examples} shows two failure cases. In \texttt{Shang\_03\_0036}, the anomaly is a person walking in circles, a subtle behavioral irregularity that generates  indistinguishable from normal pedestrian activity the VLM caption interprets the scene as routine behavior, producing a score of only 0.1. This reflects a known limitation of VLMs in distinguishing trajectory-level semantics from appearance alone, rather than a failure of MOG. In \texttt{Shang\_01\_0140}, the VLM correctly perceives a motorcycle in a pedestrian plaza yet assigns a score of only 0.2, as the LLM fails to flag it as anomalous without explicit pedestrian-only context, a limitation of zero-shot semantic grounding rather than visual recognition.

\vspace{-.5cm}
\clearpage
\section{Prompts}

The effectiveness of our motion-adaptive video anomaly detection framework relies on designed prompts that guide the Vision-Language Models (VLMs) to extract relevant information. This section details the specific prompts used at different stages of our pipeline and the rationale behind their design.

\vspace{-.5cm}
\subsection{Frame-Level Captioning Prompt}

For the frame-level analysis using Qwen/Qwen3-VL-8B-Instruct, we employ a structured prompt designed to extract comprehensive scene understanding while maintaining focus on security-relevant elements. The complete prompt is:

\begin{tcolorbox}[
  colback=gray!3!white,
  colframe=gray!60!black,
  title=Frame Captioning Prompt
]
\small
Analyze this surveillance footage frame and provide:\\
1. SCENE DESCRIPTION:\\
\ \ \ - All people present: their count, positions, activities, and interactions\\
\ \ \ - Any unusual behaviors or events\\
\ \ \ - Objects of interest: bags, weapons, vehicles, or suspicious items\\
\ \ \ - Environmental context: location type (street, pedestrian area,\\
\ \ \ \ \ parking lot, etc.), crowd density\\
\ \ \ - Vehicles in pedestrian areas: note any cars, bicycles, skateboards,\\
\ \ \ \ \ or carts in pedestrian-only zones\\
\ \ \ - Any safety or security concerns
\end{tcolorbox}

\noindent\textbf{Design Rationale.} This prompt is structured to elicit detailed, security-focused descriptions from the VLM through several key design principles:

\begin{itemize}
    \item \textbf{Explicit Entity Enumeration:} By explicitly requesting "count, positions, activities, and interactions" for people, we ensure quantitative information that is crucial for anomaly detection (e.g., crowd formation, group behaviors).
    
    \item \textbf{Open-Ended Safety Assessment:} The final request for "safety or security concerns" allows the model to identify potential anomalies that may not fit predefined categories, leveraging the VLM's general knowledge of unsafe situations.
\end{itemize}

\noindent\textbf{Prompt Considerations.}

\textit{Zero-Shot Generalization:} The prompt avoids dataset-specific terminology, enabling the same prompt to work across different surveillance scenarios (street scenes, campus footage, traffic monitoring) without modification.

\vspace{-.5cm}
\subsection{Frame-Level Anomaly Scoring Prompt}

Following the frame captioning stage, we employ Qwen/Qwen3-4B-Instruct as a specialized text-based classifier to determine whether the generated descriptions indicate normal or anomalous behavior. The system prompt is designed to provide clear classification criteria while ensuring robust, independent assessments:

\begin{tcolorbox}[
  colback=gray!3!white,
  colframe=gray!60!black,
  title=Anomaly Classification System Prompt
]
\small
You are an expert AI system specialized in video surveillance anomaly\\
detection. Analyze scene descriptions and classify them as NORMAL or ANOMALY.\\
\textbf{ANOMALY INDICATORS:}\\
- Violence: fighting, assault, weapons, threatening behavior\\
- Explosions: fire, smoke, blast effects, debris, damaged structures\\
- Accidents: car collisions, crashes, overturned vehicles\\
- Theft: shoplifting, robbery, stealing, grabbing items and fleeing\\
- Medical emergencies: injuries, falls, people on ground needing help\\
- Suspicious behavior: inappropriate running, chasing, panic, fleeing\\
- Property damage: vandalism, destruction, breaking\\
- Abandoned objects: unattended bags, suspicious packages\\
- Crowd anomalies: panic, stampede, unusual mass movements\\
- Fire or smoke visible\\
- Trespassing in restricted areas\\
\\
\textbf{NORMAL INDICATORS:}\\
- Routine activities: walking, standing, sitting, talking normally\\
- Shopping, browsing, work, recreational activities\\
- Vehicles in designated areas: roads, parking lots, bike lanes\\
- Regular traffic flow\\
- Appropriate social interactions\\
- Empty/quiet scenes with no damage or suspicious objects\\
\end{tcolorbox}

\noindent\textbf{Design Rationale.} We provide exhaustive lists of both anomalous and normal behaviors to reduce ambiguity and ensure consistent classification across diverse scenarios, rather than relying on the model's implicit understanding of "anomaly."

\noindent\textbf{Integration with Motion-Adaptive Sampling.} This frame-level classification prompt operates on selectively sampled frames identified by our TGM-based motion detection. The design ensures that when expensive VLM inference is triggered by motion, the subsequent classification is both accurate and informative, maximizing the value extracted from each processed frame.

\subsection{Temporal-Level Multi-Frame Captioning Prompt}

We employ Video-LLaMA 8B to process non-overlapping temporal windows for capturing action dynamics across multiple frames. The temporal captioning prompt is designed to elicit video-level descriptions that encode motion trajectories and sequential activities.

\begin{tcolorbox}[
  colback=gray!3!white,
  colframe=gray!60!black,
  title=Temporal Multi-Frame Captioning Prompt
]
\small
Describe the actions, movements, and events visible in this sequence.\\
\\
Specifically identify any suspicious, unusual, or anomalous behavior such as:\\
theft, shoplifting, violence, fighting, accidents, burglary, or other criminal activities.
\end{tcolorbox}

\noindent\textbf{Design Rationale.} This prompt architecture addresses the unique requirements of temporal video understanding:

\begin{itemize}    
    \item \textbf{Motion-Centric Language:} The emphasis on "actions, movements, and events" guides the model to focus on temporal dynamics rather than static scene elements, ensuring descriptions capture the behavioral evolution across frames.
    
    \item \textbf{Anomaly-Focused:} While maintaining open-ended description capabilities, the explicit listing of suspicious behaviors (theft, violence, fighting, etc.) ensures the model remains vigilant for security-relevant patterns that might be subtle or easily overlooked.

\end{itemize}

\noindent These carefully designed prompts enable our framework to extract security-relevant information at both frame and temporal levels while maintaining zero-shot generalization across diverse surveillance scenarios.

\section{Limitations}
\label{sec:limitation}
Because TGM operates on raw temporal gradients, abrupt illumination changes (e.g.,~lights switching on/off) and camera jitter can produce spurious activations that erode efficiency gains. The adaptive EMA baseline ($\alpha=0.95$) and z-score normalization absorb gradual drift but cannot distinguish a true motion burst from a global photometric spike within a single window. Our multiscale design ($\mathcal{W}=\{1,2,4\}$) partially mitigates this, a brief photometric spike elevates the $w=1$ response but is diluted at $w=2$ and $w=4$, so the aggregated signal is less likely to exceed the z-score threshold than a sustained physical motion event would.


\end{document}